\documentclass[11pt]{article}

\usepackage[sectionbib]{natbib}
\usepackage{algorithm}
\usepackage{algorithmic}
\usepackage{booktabs}
\usepackage{lineno}
\usepackage{graphicx}
\usepackage{color,xcolor}
\usepackage{subfigure} 
\usepackage{geometry}
\usepackage{multirow}

\usepackage[colorlinks=true,linkcolor=blue,citecolor=blue]{hyperref}%

\usepackage{amsmath}
\usepackage{amssymb}
\usepackage{amsthm}

\usepackage[utf8]{inputenc} 
\usepackage{hyperref}       
\usepackage{url}            
\usepackage{booktabs}       
\usepackage{amsfonts}       
\usepackage{nicefrac}       
\usepackage{microtype}      

\usepackage{natbib}

\usepackage{array,epsfig,rotating}
\usepackage{tabularx}
\usepackage{subfigure}
\usepackage{graphicx}
\usepackage{verbatim}
\usepackage{arydshln}
\usepackage{authblk}

\newtheorem{theorem}{Theorem}

\newtheorem{prop}{Proposition}

\newtheorem{assumption}{Assumption}

\allowdisplaybreaks

\def\hat{\widehat}

\newcommand{\Pb}{\mathbb{P}}

\newcommand{\E}{\mathbb{E}}

\newcommand{\Holder}{\text{H\"{o}lder}}

\newcommand{\bA}{\mathbf{A}}
\newcommand{\be}{\mathbf{e}}
\newcommand{\bV}{\mathbf{V}}
\newcommand{\bv}{\mathbf{v}}
\newcommand{\bD}{\mathbf{D}}
\newcommand{\bX}{\mathbf{X}}
\newcommand{\bx}{\mathbf{x}}
\newcommand{\bZ}{\mathbf{Z}}
\newcommand{\bz}{\mathbf{z}}
\newcommand{\bg}{\mathbf{g}}

\newcommand{\bS}{\mathbf{S}}

\newcommand{\bbeta}{\boldsymbol{\beta}}

\newcommand\independent{\protect\mathpalette{\protect\independenT}{\perp}}
\def\independenT#1#2{\mathrel{\rlap{$#1#2$}\mkern2mu{#1#2}}}

\DeclareMathOperator*{\argmin}{argmin}

\def\spacingset#1{\renewcommand{\baselinestretch}%
{#1}\small\normalsize} \spacingset{1}
\spacingset{1}

\usepackage{float}

\begin{document}



\title{Continuous Treatment Effects with Surrogate Outcomes
}
\author{Zhenghao Zeng$^1$, David Arbour$^2$, Avi Feller$^3$,  \\ 
 Raghavendra Addanki$^2$, Ryan Rossi$^2$, Ritwik Sinha$^2$, Edward H. Kennedy$^1$\\ $\quad$ \\
    $^1$Department of Statistics and Data Science \\
    Carnegie Mellon University \\ 
    $^2$Adobe Research \\
    $^3$Goldman School of Public Policy and Department of Statistics \\
    University of California, Berkeley
    }

  \date{}

\maketitle

\begin{abstract}
In many real-world causal inference applications, the primary outcomes (labels) are often partially missing, especially if they are expensive or difficult to collect. If the missingness depends on covariates (i.e., missingness is not completely at random), analyses based on fully observed samples alone may be biased. Incorporating surrogates, which are fully observed post-treatment variables related to the primary outcome, can improve estimation in this case. In this paper, we study the role of surrogates in estimating continuous treatment effects and propose a doubly robust method to efficiently incorporate surrogates in the analysis, which uses both labeled and unlabeled data and does not suffer from the above selection bias problem. Importantly, we establish the asymptotic normality of the proposed estimator and show possible improvements on the variance compared with methods that solely use labeled data. Extensive simulations show our methods enjoy appealing empirical performance. 
\end{abstract}


\def\spacingset#1{\renewcommand{\baselinestretch}%
{#1}\small\normalsize} \spacingset{1}
\spacingset{1}

\section{Introduction}
In many causal inference applications, the primary outcomes are missing for a non-trivial number of observations.
For instance, in studies on long-term health effects of medical interventions, some measurements require expensive testing and a loss to follow-up is common \citep{hogan2004handling}. In evaluating commercial online ad effectiveness, some individuals may drop out from the panel because they use multiple devices \citep{shankar2023direct}, leading to missing revenue measures. 
In many of these studies, however, there often exist short-term outcomes that are easier and faster to measure, e.g., short-term health measures or an online ad's click-through rate, that are observed for a greater share of the sample. These outcomes, which are typically informative about the primary outcomes themselves, are referred to as \textit{surrogate outcomes} or \textit{surrogates}. 

There is a rich causal inference literature addressing missing outcome data. Simply restricting to data with observed primary outcomes may induce strong bias \citep{hernan2010causal}. 
Ignoring unlabeled data also reduces the effective sample size for estimating the treatment effects and inflates the variance. \citet{chakrabortty2022general} considered the missing completely at random (MCAR) setting and showed that incorporating unlabeled data reduces variance. \citet{zhang2023semi} generalized the results to missing-at-random (MAR) settings where the unlabeled data has a much larger size than the labeled data. \citet{kallus2020role} further examined the role of surrogates in datasets with limited primary outcomes and showed efficiency gains after including surrogates and unlabeled data in the analysis. \citet{singh2022generalized} proposed a generalized kernel ridge regression framework, which incorporates information on surrogate outcomes in treatment effect estimation. See also \citet{zhang2022high, hou2021efficient, zeng2023efficient} for relevant discussions. 

Continuous treatments appear in many applications; e.g., waiting time before follow-up, percent of discount, and drug dosage. Existing estimation procedures include outcome modeling \citep{newey1994kernel} and treatment process modeling \citep{galvao2015uniformly}. \citet{kennedy2017non} proposed doubly robust methods that model both the outcome and the treatment process, and enjoy appealing robustness properties. \citet{bonvini2022fast} further examined high-order estimators to achieve faster rates. 

In this paper, we consider the estimation of dose-response functions with limited primary outcome data. We propose novel doubly robust methods using both labeled and unlabeled data, with the help of surrogates. Our approach avoids the potential selection bias caused by restricting only to labeled data and provably reduces the variance. Importantly, we study the theoretical properties of the estimator proposed, and establish its asymptotic normality to facilitate statistical inference. Our work serves as a counterpart for \citet{kallus2020role} in continuous treatment effects settings and enriches the existing dose-response estimation literature \citep{kennedy2017non, bonvini2022fast}.

The rest of this paper is organized as follows: In Section \ref{sec:setup} we introduce the problem setup and notation. Section \ref{sec:identification} provides assumptions to identify the continuous treatment effect. Novel methodology and theoretical guarantees are discussed in Section \ref{sec:DR-estimation}. Our simulation study in Section \ref{sec:simulation} demonstrates the performance of our method. Finally we conclude with a discussion in Section \ref{sec:discussion}. All proofs, additional simulation results and a real data example are included in supplementary materials.

\section{Setup and Notation}\label{sec:setup}

In this section we introduce the problem of estimating continuous treatment effects with surrogates. We formalize the problem using potential outcomes framework \citep{splawa1990application, rubin1974estimating} and introduce notation to present our results concisely.

\subsection{Data Structure}

Suppose we have access to two datasets $\mathcal{L}$ and $\mathcal{U}$ from a randomized experiment/observational study. The labeled dataset is $\mathcal{L} = \{ \bZ_i = (\bV_i, \bS_i, A_i,Y_i,R_i=1), 1\leq i \leq n_1 \}$, where $\bV$ is the set of covariates, $\bS$ is the vector of surrogates, $A$ is the continuous treatment, and $Y$ is the primary outcome. Here $R$ is an indicator with $R=1$ if a sample is from the labeled data $\mathcal{L}$ and $R=0$ otherwise. The unlabeled dataset consists of samples without primary outcomes $\mathcal{U} = \{\bZ_i = (\bV_i, \bS_i, A_i, R_i = 0), n_1+1 \leq i \leq n_1 + n_2 \}$. Hence the total sample size is $n=n_1 + n_2$ and both $\mathcal{L}$ and $\mathcal{U}$ contain information on treatment $A$, surrogates $\bS$ and covariates $\bV$, but the primary outcome is missing in the unlabeled dataset $\mathcal{U}$ due to, for example, loss to follow-up. Let $\bX = (\bV, \bS)$ be the union of covariates and surrogates. Note that we will present the results for $\bS \neq \emptyset$, but $\bS = \emptyset$ can be viewed as a special setting where our methodology still applies.

\subsection{Estimand and Nuisance Functions}
Now we define the continuous treatment effects of interest. We use the random variable $Y^a$ to denote the potential (counterfactual) outcome we would have observed
had a subject received treatment $A = a$, which may be contrary to the observed $Y$. The continuous treatment effect (or dose-response function) is defined as
\begin{equation}\label{eq:dose-response}
    \theta(a) = \E[Y^a].
\end{equation}
Without additional causal assumptions introduced in Section \ref{sec:identification}, $\theta(a)$ involves counterfactual outcome $Y^a$ and cannot be identified by observed data. Note that since $\bS$ contains post-treatment surrogate outcomes and may be affected by the treatment, we also use potential outcomes $\bS^a$ to denote the surrogate under treatment $A=a$.

We further introduce some nuisance functions that are not of primary interest, but which our estimation method depends on. Let $\mu(A, \bX, 1) = \E[Y|A,\bX,R=1]$ be the regression function of the primary outcome in the labeled population $R=1$. Denote the function obtained by further regressing $\mu(A,\bX, 1)$ on $(A,\bV)$ as $\tau(A,\bV) = \E[\mu(A,\bX,  1)|A,\bV]$ , where the expectation is over the conditional distribution of $\bS$ given $A,\bV$. Denote the conditional density of $A$ given $\bV$ as $\pi(a|\bV)$ and the marginal density of $A$ as $f(a) = \int \pi(a|\bv) d\Pb(\bv)$. The ratio of the marginal density and conditional density is denoted as $w(a, \bv) = f(a)/\pi(a|\bv)$. $w(a,\bv)$ is known as a stabilized weight in the literature \citep{robins2000marginal, ai2021unified}. The propensity score for $R$ (i.e. the conditional probability that the primary outcome is observed) is denoted as $\rho(A,\bX) = \Pb(R=1|A,\bX)$. 

Let $(\mathcal{V}, \mathcal{S}, \mathcal{A})$ denote the support of $(\bV, \bS, A)$. For a (possibly random) function $f$ on variables $\bZ$ we use $\Pb_n [f(\bZ)] $ or $\Pb_n [f]$ to denote the sample average $\frac{1}{n}\sum_{i=1}^n f(\bZ_i)$ on a sample of size $n$. The sample over which averages are taken should be clear from context. We use $\Pb[f] = \int f(\bz) d \Pb(\bz)$ to denote the expectation of $f(\bZ)$ where only randomness of $\bZ$ is considered and $f$ is conditioned on. Finally we use $\|f\|_{\infty} = \sup_{\bz \in \mathcal{Z}} |f(\bz)|$ to denote the uniform norm, $\|f\|_a = \left\{\int f^2(\bz) d \Pb(\bz|A=a)\right\}^{1/2}$ to denote the $L_2$-norm with respect to the conditional distribution $\bZ |A=a$ and $\|f\|_2 = \left\{\int f^2(\bz) d \Pb(\bz)\right\}^{1/2}$ to denote the usual $L_2$-norm. 

\section{Identification}\label{sec:identification}
In this section we discuss sufficient conditions to identify the dose-response function \eqref{eq:dose-response}, summarized as follows:
\begin{assumption}\label{asm:consistency}
    (Consistency) $Y=Y^a, \bS = \bS^a$ if $A=a, a \in \mathcal{A}$.
\end{assumption}
\begin{assumption}\label{asm:surrogates-A}
    (Exchangeability) $(Y^a, \bS^a) \independent A \mid \bV$ for $a \in \mathcal{A}$.
\end{assumption}
\begin{assumption}\label{asm:surrogates-R}
    (Missing at random) $R \independent Y^a \mid \bV, \bS^a, A=a$ for $a \in \mathcal{A}$.
\end{assumption}
\begin{assumption}\label{asm:surrogates-positivity}
    (Positivity) $\pi(a|\bV) > 0,  \rho(a,\bX)>0 $ almost surely for $a \in \mathcal{A}$. 
\end{assumption}
\begin{figure}
    \centering
    \includegraphics[width=2.5in]{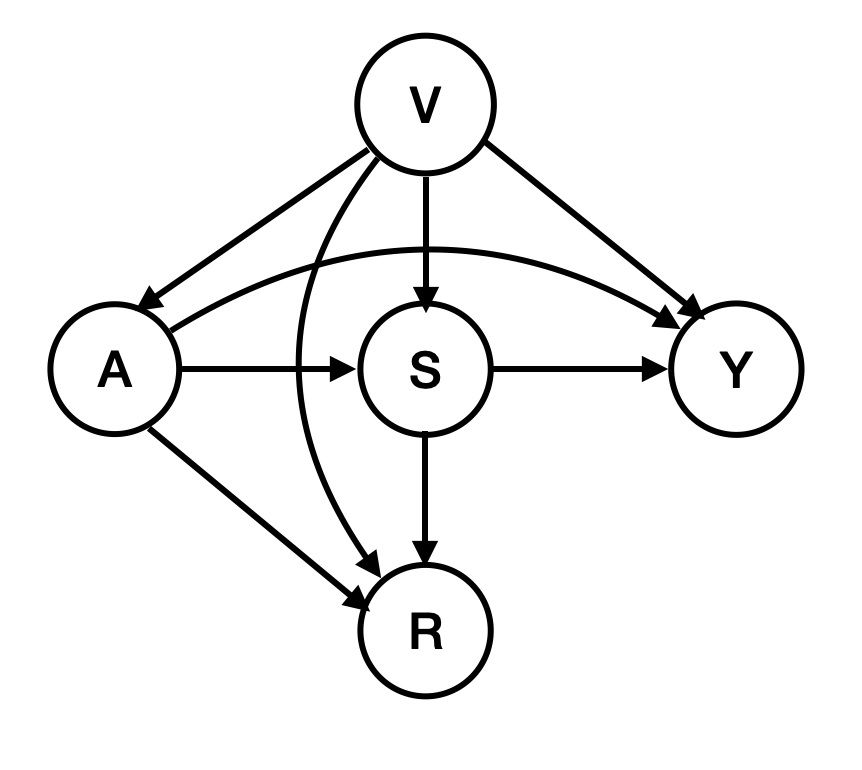}
    \caption{Example of a causal graph with surrogate outcome $\bS$.}
    \label{fig:DAG}
\end{figure}
Assumption \ref{asm:consistency} is also known as the stable unit treatment value assumption (SUTVA), and requires an absence of interference between different individuals in the study. Assumption \ref{asm:surrogates-A} is commonly used to identify treatment effects and holds in a randomized experiment or observational study with all confounders measured. Assumption \ref{asm:surrogates-R} ensures whether the primary outcome is observed only depends on covariates $\bV$, surrogates $\bS$, and treatment $A$ so that the distributions of labeled and unlabeled data are comparable after conditioning on $\bX,A$. Assumption \ref{asm:surrogates-positivity} says every subject has some chance to receive treatment $A=a$ and has the primary outcome observed. An illustrative causal graph is shown in Figure \ref{fig:DAG}. The readers are referred to \citet{gill2001causal} for detailed discussions on identifying continuous treatment effects and \citet{kallus2020role} for the role of surrogates in identifying treatment effects. With all these assumptions, the treatment effects of interest can be identified using the observable distribution as summarized in Theorem \ref{thm:identification}.

\begin{theorem}\label{thm:identification}
    Under Assumption \ref{asm:consistency}--\ref{asm:surrogates-positivity} we have
    \begin{equation}\label{eq:identification}
    \begin{aligned}
    \theta(a)  =&\, \E \{ \E[ \E(Y|A=a,\bX,R=1)|A=a, \bV] \} \\
    = &\, \E \left\{ \E[\mu(a,\bX,1)|A=a,\bV] \right\} \\
    = &\, \E[\tau(a,\bV)]
    \end{aligned}
    \end{equation}
     for fixed $a \in \mathcal{A}$, where the expectations are over $Y, \bS, \bV$ in \eqref{eq:identification}.
\end{theorem}
The identification formula \eqref{eq:identification} suggests the following plug-in style estimator of $\theta(a)$: first regress $Y$ on $A, \bX$ in the labeled dataset $\mathcal{L}$ and obtain an estimator of $\mu$ as $\widehat{\mu}$, which is further regressed on $A, \bV$ to get an estimator $\widehat{\tau}$. Finally we take an average over all the samples to get an estimator of $\theta(a)$ as 
\begin{equation}\label{eq:plugin}
    \widehat{\theta}(a) = \Pb_n [\widehat{\tau}(a,\bV)].
\end{equation}
The performance of such a plug-in style estimator highly depends on the estimation error of $\widehat{\tau}$. To see this, assume the nuisance estimator $\widehat{\tau}$ is independent of the samples that we average over, then the conditional bias of $\widehat{\theta}(a)$ given $\widehat{\tau}$ is
\[
\E\left[\widehat{\theta}(a)\right]-\theta(a) = \E\left[\widehat{\tau}(a,\bV) - \tau(a,\bV)\right],
\]
which solely depends on the estimation error of $\tau$. When $\tau$ is hard to estimate (e.g., non-smooth/sparse in high-dimensional problems), the plug-in style estimator may inherit the slow convergence rate of $\widehat{\tau}$ and have sub-optimal performance.

\section{Doubly Robust Estimation}\label{sec:DR-estimation}

In this section we present the main results of this paper. We begin with an alternative characterization of the dose-response function \eqref{eq:dose-response} in Section \ref{sec:DR-characterization}, based on which we propose a doubly robust estimator in Section \ref{sec:DR-procedure}. Finally, theoretical guarantees of the proposed method are provided in Section \ref{sec:DR-theory}.  

\subsection{Doubly Robust Characterization}\label{sec:DR-characterization}

Since treatment $A$ is continuous, the function $\theta(a)$ in \eqref{eq:identification} is not pathwise differentiable \citep{bickel1993efficient, diaz2013targeted} and we need a novel way to apply semiparametric efficiency theory. The idea is to find a pseudo-outcome $\varphi(\bZ):=\varphi(\bZ;\mu, \pi, \rho)$ depending on nuisance functions such that $\E[\varphi(\bZ;{\mu}, {\pi}, {\rho})| A=a] = \theta(a)$, i.e., regressing $\varphi(\bZ;\mu, \pi, \rho)$ on $A$ yields the target dose-response function ideally with second-order dependence on nuisance estimation error. Following \citet{rubin2005general, kennedy2017non}, we consider the functional $\psi = \E[\theta(A)]$, which is pathwise differentiable and admits an efficient influence function. Then the pseudo-outcome $\varphi(\bZ;\mu, \pi, \rho)$ is a component in the influence function of $\psi$. We omit the derivation of the influence function and only present the form of pseudo-outcome. Let $(\bar{\mu}, \bar{\pi}, \bar{\rho})$ be nuisance functions that may not necessarily equal the true $({\mu}, {\pi}, {\rho})$, and define
\[
\varphi(\bZ; \bar{\mu}, \bar{\pi}, \bar{\rho}) =\left[\frac{R(Y-\bar{\mu}(A,\bX,1))}{\bar{\rho}(A,\bX)} + \bar{\mu}(A,\bX,  1 ) - \bar{\tau}(A,\bV) \right]\frac{\int_{\mathcal{V}} \bar{\pi}(A|\bv) d \Pb(\bv)}{\bar{\pi}(A|\bV)} + \int_{\mathcal{V}} \bar{\tau}(A,\bv) d \Pb(\bv)
\]
where $\bar{\tau}(A,\bV) = \E[\bar{\mu}(A,\bX,1)|A,\bV]$. The following theorem shows an alternative characterization of the dose-response function $\theta(a)$ through $\varphi(\bZ; \bar{\mu}, \bar{\pi}, \bar{\rho})$.

\begin{prop}
    \label{prop:surrogates-dr}
    Let $(\bar{\mu}, \bar{\pi}, \bar{\rho})$ be nuisance functions that may not necessarily equal the true $({\mu}, {\pi}, {\rho})$. Then
    \[
    \E[\varphi(\bZ;\bar{\mu}, \bar{\pi}, \bar{\rho})| A=a] = \theta(a)
    \]
    if either $\bar{\mu} = \mu$ or $(\bar{\pi}, \bar{\rho}) = (\pi, \rho)$.
\end{prop}

Proposition \ref{prop:surrogates-dr} gives the first interpretation of the double robustness of our methods. There are two chances to obtain the dose-response function $\theta(a)$ by regressing the pseudo-outcome $\varphi(\bZ;\bar{\mu}, \bar{\pi}, \bar{\rho})$ on $A$: we correctly specify either the outcome regression model $\mu$ or both the propensity score of $R$ and conditional density of $A$ given $\bV$ (although see Proposition \ref{prop:surrogates-bias} for a slightly different parameterization).  

\subsection{Estimation Procedure}\label{sec:DR-procedure}
The doubly robust characterization in Proposition \ref{prop:surrogates-dr} motivates a two-stage procedure to estimate $\theta(a)$: In the first step we model the nuisance functions that appear in the pseudo-outcome $\varphi(\bZ)$ with flexible nonparametric or machine learning methods. We then construct estimated pseudo-outcomes $\widehat{\varphi}(\bZ)$ and regress these on $A$ to obtain an estimator of $\theta(a)$. We will use the stability framework developed in \citet{kennedy2023towards} to analyze such an estimator, which regresses imputed outcomes $\widehat{\varphi}(\bZ)$ on treatment. The formal procedure is summarized in Algorithm \ref{alg:DR-surrogates}.

\begin{algorithm}[tb]
\caption{Doubly Robust Estimation}
\label{alg:DR-surrogates}
    Let $(D_1^n, D_2^n, T^n)$ denote three independent samples of $n$ i.i.d observations of $\bZ$ and $D^n = (D_1^n, D_2^n)$ denote the training set to train the nuisance functions.
    \begin{enumerate}
        \item Nuisance functions training: Construct estimates of $\mu, \tau, \rho, w$ using $D_1^n$. Then use $D_2^n$ to get an initial estimator of $\theta(a)$ as
        \[
        \widehat{\theta}_0(a) = \frac{1}{n} \sum_{i \in D_2^n} \widehat{\tau}(a, \bV_i).
        \]
        \item Pseudo-outcome regression: Construct estimated pseudo-outcome
        \[
        \widehat{\varphi}(\bZ)=\left[\frac{R(Y-\widehat{\mu}(A,\bX,1))}{\widehat{\rho}(A,\bX)} + \widehat{\mu}(A, \bX, 1) - \widehat{\tau}(A, \bV) \right]\widehat{w}(A,\bV) + \widehat{\theta}_0(A)
        \]
        for each sample in $T^n$ and regress the pseudo-outcomes on the treatment $A$ in $T^n$ using a linear smoother to obtain 
        \[
        \widehat{\theta}(a) = \widehat{\E}_n\left[\widehat{\varphi}(\bZ)|A=a\right] = \sum_{i \in T^n} W_i(a;\bA^n) \widehat{\varphi}(\bZ_i),
        \]
        where $\bA^n = (A_1, \dots, A_n)$ are the treatments in $T^n$ and $W_i(a;\bA^n)$ is the coefficient of $i$-th sample.

        \item (Optional) Cross-fitting: Swap the role of $D_1^n, D_2^n, T^n$ and repeat steps 1 and 2. Use the average of different estimates as the final estimator of $\theta(a)$.
    \end{enumerate}
\end{algorithm}

Sample splitting is used in Algorithm \ref{alg:DR-surrogates} to avoid complicated empirical process assumptions that are difficult to justify in practice and simplify our theoretical analysis \citep{robins2008higher, chernozhukov2018double, kennedy2020sharp, levis2023covariate, bonvini2023flexibly}. In Step 1, one can use any appropriate regression/classification algorithms to estimate $\mu, \tau, \rho$. However, there are fewer results on estimating stabilized weight $w(a,\bv) = f(a)/\pi(a|\bv)$. \citet{ai2021unified} proposed a method that directly estimates $w$ with entropy maximization. Alternatively, one can estimate the conditional density $\pi$ \citep[see, e.g.,][for discussions on related methods]{colangelo2020double}, and then the marginal density can be estimated by
\[
\widehat{f}(a) = \frac{1}{n} \sum_{i \in D_2^n} \widehat{\pi}(a| \bV_i)
\]
and use their ratio to estimate $w$. 

Our analysis in Section \ref{sec:DR-theory} applies to a wide class of nuisance estimators if the product of convergence rates is fast enough. In step 2 we focus on linear smoother-based estimators since they are relatively straightforward to analyze. We believe similar results hold for a wider class of regression estimators under the stability framework in \citet{kennedy2023towards} and leave the theoretical analysis of applying general regression algorithms for future investigation. In applications, researchers can choose suitable parametric methods based on their domain knowledge or flexible nonparametric machine learning methods to avoid model misspecification (or ensembles thereof). 

\subsection{Theoretical Results}\label{sec:DR-theory}

We first present estimation error guarantees of Algorithm \ref{alg:DR-surrogates}, for general linear smoothers. Then we focus on a specific type of linear smoother, namely local linear regression, and study the asymptotic distribution of the estimator in Section \ref{sec:DR-normality}.

\subsubsection{Oracle Estimation Theory}\label{sec:DR-oracle}

In Algorithm \ref{alg:DR-surrogates} we obtain the estimator $\widehat{\theta}(a)$ by regressing the imputed pseudo-outcome $\widehat{\varphi}(\bZ)$ on $A$. It is natural to compare $\widehat{\theta}(a)$ with the ``oracle" estimator $\widetilde{\theta}(a)$ defined as
\[
\widetilde{\theta}(a) = \sum_{i \in T^n}W_i(a;\bA^n)\varphi(\bZ_i),
\]
which regresses the ground-truth pseudo-outcome $\varphi(\bZ)$ on $A$ using the same linear smoother. Intuitively it is hard for $\widehat{\theta}$ to have a faster convergence rate than $\widetilde{\theta}$. In the following theorem, we summarize the conditions under which $\widehat{\theta}$ enjoys the same rate as $\widetilde{\theta}$ and hence is ``oracle efficient".

\begin{theorem}\label{thm:DR-surrogates}
    Let $\widehat{\theta}(a)$ denote the doubly robust estimator obtained from Algorithm \ref{alg:DR-surrogates} and $\widetilde{\theta}(a) = \sum_{i \in T^n}W_i(a;\bA^n)\varphi(\bZ_i)$ denote the oracle estimator with oracle risk $R_n^2(a) = \E[(\widetilde{\theta}(a)-\theta(a))^2]$ at point $a$. Suppose 
    \begin{itemize}
        \item $\operatorname{Var}(\varphi(\bZ)|A=a)\geq c,  \; \forall\, a \in \mathcal{A}$ for some constants $c>0$.
        \item The estimator for nuisance functions in Step 2 is uniformly consistent in the sense that $\sup_{\bz}|\widehat{\varphi}(\bz) - \varphi(\bz)| = o_{\Pb}(1)$.
    \end{itemize} 
    Then we have
    \[
    \widehat{\theta}(a) - \theta(a) = \widetilde{\theta}(a) - \theta(a) + \widehat{\E}_n\left[\widehat{b}(A)|A=a\right] + o_{\Pb}\left(R_n(a)\right)
    \]
    where $\widehat{b}(a) = \E[\widehat{\varphi}(\bZ) - \varphi(\bZ)|D^n, A=a]$ is the conditional bias of $\widehat{\varphi}(\bZ)$ and $\widehat{\E}_n\left[\widehat{b}(A)|A=a\right]=\sum_{i \in T^n}W_i(a;\bA^n)\widehat{b}(A_i)$. Further assume
    \[
    \widehat{\E}_n\left[\widehat{b}(A)|A=a\right] = o_{\Pb}(R_n(a)),
    \]
    then $\widehat{\theta}$ is oracle efficient in the sense that
    \[
    \frac{\widehat{\theta}(a) - \widetilde{\theta}(a)}{R_n(a)} \overset{P}{\rightarrow} 0.
    \]    
\end{theorem}

The conditions in Theorem \ref{thm:DR-surrogates} are mild: $\varphi(\bZ)$ is not a function of $A$ and hence its conditional variance given $A$ should be positive; We do not impose conditions on the convergence rate of $\widehat{\varphi}$ but only require its consistency. The key condition for $\widehat{\theta}$ to be oracle efficient is $\widehat{\E}_n\left[\widehat{b}(A)|A=a\right] = o_{\Pb}(R_n(a))$ and hence we need a bound on the conditional bias $\widehat{b}(a)$, as summarized in the following proposition.

\begin{prop}
    \label{prop:surrogates-bias}
    Under the conditions in Theorem \ref{thm:DR-surrogates}, further assume the estimated conditional probability $\widehat{\rho}(a,\bx) \geq c$ and the estimated stabilized weight $\widehat{w}(a,\bv) \leq C$ hold for all $ a \in \mathcal{A}, \bx \in \mathcal{X}, \bv \in \mathcal{V} $ for some constant $c, C>0$.
    Then we have
    \[
    |\widehat{b}(a)| \lesssim \|\widehat{\rho}-\rho\|_a \|\widehat{\mu}-\mu\|_a + \|\widehat{\tau}-\tau\|_a \|\widehat{w}-w\|_a+ \frac{1}{n} \sum_{i \in D_2^n} \widehat{\tau}(a,\bV_i) - \Pb[\widehat{\tau}(a, \bV)]
    \]
    where recall $\|f\|_a^2 = \int f^2(\bz) d \Pb(\bz|A=a)$. If we further assume the weights of the linear smoother satisfy $\sum_{i \in T^n} W_i(a;\bA^n) \leq C$ and there exists a neighborhood $N(a)$ around $a$ such that $W_i(a;\bA^n)=0$ if $A_i \notin N(a)$. Then we have
    \[
    \widehat{\E}_n\left[\widehat{b}(A)|A=a\right] \lesssim \sup_{t \in N(a)}| \widehat{b}(t)|.
    \]
\end{prop}
We note that the condition on linear smoothers will be satisfied by Nadaraya–Watson estimators \citep{nadaraya1964estimating, watson1964smooth} and local polynomial estimators \citep{fan1994robust, fan1996local} when the kernel function used has compact support, e.g., the uniform and Epanechnikov kernel. In the bound for $\widehat{b}(a)$, the first two terms depend on the product of the convergence rates of nuisance functions. This phenomenon is commonly observed in influence functions-based doubly robust approaches \citep{kennedy2016doublerobust, chernozhukov2018double, meza2021nested} and gives the second interpretation of double robustness: Introducing an extra term to the plug-in style estimator in $\varphi$ can correct for the first-order bias, making the remainder second-order and ``doubly small". In common examples of ATE estimation, conditional bias only involves estimation error of outcome model and propensity scores. The additional dependence on the convergence rate of $\hat{\tau}$ (compared with Proposition \ref{prop:surrogates-dr}) appears in the bias since the formula \eqref{eq:identification} is more complicated compared with ATE-style functional and we use an agnostic estimator of $\tau$ in Algorithm \ref{alg:DR-surrogates}. In most settings we expect the nuisance error rate $\|\widehat{\mu}-\mu\|_a$ (with respect to measure $d \Pb(\bz|A=a)$) has the same order as the more common conventional rate $\|\widehat{\mu}-\mu\|$ (with respect to measure $d \Pb(\bz)$), which is $n^{-\alpha/(2\alpha+d+1)}$ if $\mu(a,\bx)$ belongs to a \Holder \, class of order $\alpha$ and $\bX$ is $d$-dimensional. Alternatively one can always upper bound $\|\widehat{\mu}-\mu\|_{a}$ with $ \|\widehat{\mu}-\mu\|_{\infty}$ at the cost of log factors \citep[Section 1.6.2]{tsybakov2009}. The empirical process term $ \frac{1}{n} \sum_{i \in D_2^n} \widehat{\tau}(a,\bV_i) - \Pb[\widehat{\tau}(a, \bV)]$ would be $O_{\Pb}(1/\sqrt{n})$ provided that $\E[\widehat{\tau}^2(a,\bV)|D^n]$ is bounded. The bound on $ \widehat{\E}_n\left[\widehat{b}(A)|A=a\right]$ in Proposition \ref{prop:surrogates-bias} involves 
\[
\sup_{t \in N(a)} \left | \frac{1}{n}\sum_{i \in D_2^n} \widehat{\tau}(t,\bV_i) - \Pb[\widehat{\tau}(t,\bV)] \right|
\]
as a coarse bound and for specific estimators, a tighter bound can be derived. For instance, as we will see in local linear estimation this empirical process term is $o_{\Pb}\left(1/\sqrt{nh} \right)$ and asymptotically negligible. Hence we can focus on the first two second-order terms in the conditional bias. Theorem \ref{thm:DR-surrogates} together with Proposition \ref{prop:surrogates-bias} gives conditions under which the estimator $\widehat{\theta}(a)$ has the same rate as the oracle estimator $\widetilde{\theta}(a)$. For instance, assume $\bV = \bX$ (no surrogates) and $|\bX| = d$, suppose the dose-response function $\theta(a)$ is $\alpha$-smooth (i.e. belongs to a \Holder \, class of order $\alpha$) and $w$ and $\mu$ are $s$-smooth, then the rate condition for $\widehat{\theta}(a)$ to be oracle efficient is
\[
n^{-\frac{2s}{2s+d+1}} \leq n^{-\frac{\alpha}{2\alpha+1}}
\]
or equivalently
\[
s \geq \frac{\alpha(d+1)}{2(\alpha+1)}.
\]

\subsubsection{Asymptotic Normality}\label{sec:DR-normality}

In the following discussions, we will analyze the estimator $\widehat{
\theta}(a)$ based on a particular linear smoother (i.e. the local linear regression estimator). For a scalar bandwidth parapmeter $h$, let $\bg_{ha}(t) = [1, (t-a)/h]^{\top}$ be the local linear basis, $K_{ha}(t) = h^{-1}K((t-a)/h)$ with $K$ being a probability density. The local linear regression solves the following weighted least square problem:
\[
\min_{\bbeta \in \mathbb{R}^2} \sum_{i \in T^n} K_{ha}(A_i) \left[ \widehat{\varphi}(\bZ_i) -\bg_{ha}(A_i)^{\top}\bbeta \right]^2
\]
which gives the closed-form solution:
\[
\widehat{\bbeta}_h(a) = \widehat{\bD}_{ha}^{-1} \Pb_n [\bg_{ha}(A)K_{ha}(A)\widehat{\varphi}(\bZ)],
\]
where $\widehat{\bD}_{ha} = \Pb_n[\bg_{ha}(A)K_{ha}(A)\bg_{ha}^{\top}(A)]$ and $\Pb_n$ is the sample average over $T^n$. Then the local linear estimator of $\theta(a)$ is $\be_1^{\top}\widehat{\bbeta}_h(a)$, i.e. the first component of $\widehat{\bbeta}_h(a)$. We summarize the asymptotic properties of this local linear estimator in Theorem \ref{thm:surrogates-normality}, which is the key contribution of our paper.

\begin{theorem}\label{thm:surrogates-normality}
    (Asymptotic normality of Local Linear Estimator) Let $a \in \mathcal{A}$ be an inner point of the compact support $\mathcal{A}$ of $A$. Assume
    \begin{enumerate}
        \item The bandwidth $h$ satisfies $h=h_n \rightarrow 0$ and $nh \rightarrow \infty$ as $n \rightarrow \infty$.
        \item The marginal density of the treatment $f(a)$ is continuously differentiable, the conditional variance $ \operatorname{Var}(\varphi(\bZ)|A=a)$ is continuous and the dose response $\theta(a)$ is twice continuously differentiable.
        \item For any $a\in \mathcal{A}, \bx \in \mathcal{X}, \bv \in \mathcal{V}$, the conditional variance $ \operatorname{Var}(\varphi(\bZ)|A=a)>0,$ the marginal density of treatment $  f(a) \geq c,$ the estimated conditional probability $ \widehat{\rho}(a,\bx) \geq c$ and the estimated stablized weight $ \widehat{w}(a,\bv) \leq C$ for some constant $C, c>0$.
        \item $K$ is a continuous symmetric probability density with support $[-1,1]$.
        \item All nuisance functions are estimated consistently in $\ell_{\infty}$ norm and the estimated pseudo-outcome also satisfies
        \[
        \|\widehat{\varphi} - \varphi\|_{\infty} = o_{\Pb}(1).
        \]
        Furthermore, the convergence rates of nuisance estimation satisfy
        \[
        \begin{aligned}
            \sup_{|t-a|\leq h} \|\widehat{\rho} - \rho\|_t \|\widehat{\mu} - \mu\|_t =&\, o_{\Pb} \left(\frac{1}{\sqrt{nh}} \right) \\
            \sup_{|t-a|\leq h} \|\widehat{\tau} - \tau\|_t \|\widehat{w} - w\|_t =&\, o_{\Pb}\left(\frac{1}{\sqrt{nh}} \right).
        \end{aligned}
        \]
    \end{enumerate}
    Then we have
    \[
    \sqrt{nh} \left( \widehat{\theta}(a) - \theta(a)-\frac{h^2 \theta''(a) \int u^2 K(u)du}{2} \right)\overset{d}{\rightarrow} N \left( 0, \frac{\sigma^2(a) \int K^2(u) du}{f(a)} \right)
    \]
    where 
    \[
    \sigma^2(a)=\E \left\{ \left[ \frac{\operatorname{Var}(Y|A=a,\bX,R=1)}{\rho(a,\bX)}+ \operatorname{Var}(\mu(a,\bX,1)|A=a, \bV) \right]w^2(a,\bV) \bigg | A=a  \right\}
    \]
\end{theorem}
Assumptions 1-4 in Theorem \ref{thm:surrogates-normality} are standard in the nonparametric kernel regression literature. Assumption 5 guarantees that the contribution of nuisance estimation error is asymptotically negligible compared with the smoothing error. One can also use a symmetric kernel $K$ supported on $\mathbb{R}$ that is square-integrable and has a finite second-order moment. Then the rate condition would be $\sup_{t \in \mathcal{A}} \|\widehat{\rho} - \rho\|_t \|\widehat{\mu} - \mu\|_t = o_{\Pb} \left(\frac{1}{\sqrt{nh}} \right)$. Similar to our discussions in Section \ref{sec:DR-oracle}, the rate conditions in assumption 5 are imposed on the product of nuisance estimation error since we use a doubly robust estimator and the conditional bias is ``second-order small". The theoretically optimal bandwidth to estimate a twice continuously differentiable function is $h \asymp n^{-1/5}$ and yields a root mean square error of order $n^{-2/5}$. With such a choice of $h$ the requirement on the convergence rate becomes $\sup_{t \in \mathcal{A}} \|\widehat{\rho} - \rho\|_t \|\widehat{\mu} - \mu\|_t = o_{\Pb} \left(n^{-2/5} \right)$, which, for example, can be satisfied when $\rho$ is consistent and $\mu$ is estimated with rate $O_{\Pb}(n^{-2/5})$. In applications, one can select the bandwidth using leave-one-out cross-validation \citep{hardle1988far} due to its computational ease. Specifically, after we obtain the estimated pseudo-outcome, we treat them as known and select $h$ by
\begin{equation}\label{eq:loocv}
    \widehat{h}_{opt} = \argmin_{h \in \mathcal{H}} \sum_{i } \left\{ \frac{\widehat{\varphi}(\bZ_i)-\widehat{\theta}_h(A_i)}{1-\widehat{W}_h(A_i)} \right\}^2.
\end{equation}
where $\widehat{W}_h(A_i) = \be_1^{\top} \widehat{\bD}_{hA_i}^{-1}\be_1 h^{-1} K(0)$. In the setting of Algorithm $\ref{alg:DR-surrogates}$ we can select the bandwidth on $D_2^n$ to avoid overfitting on $T^n$.

Similar to most nonparametric inference methods, Theorem \ref{thm:surrogates-normality} shows the estimator $\widehat{\theta}$ is centered around $\bar{\theta}(a) = \theta(a) - h^2 \theta''(a) \int u^2 K(u)du /2$ instead of $\theta(a)$ under optimal smoothing, which is known as the ``bias problem" in the literature \citep[Section 5.7]{wasserman2006all}. 
There are several methods to overcome the bias problem and each of them has its own consideration and trade-offs. For instance, one can estimate the second-order derivative and debias the estimator \citep{calonico2018effect, takatsu2022debiased} but this requires extra smoothness conditions. Another method is to undersmooth \citep{fan2022estimation} and make the bias decrease asymptotically relative to the variance. Unfortunately, there does not seem to be a simple, practical rule for choosing just the right amount of undersmoothing. In this paper we choose to live with the bias and report uncertainty quantification for $\bar{\theta}$. Theoretically, the bias shrinks to 0 as $n \rightarrow \infty$ and the proposed estimator $\hat{\theta}(a)$ is still consistent for $\theta(a)$. To construct confidence intervals for $\bar{\theta}(a)$ one needs to estimate the variance. Define a localized functional $\theta_h(a) = \be_1^{\top} \bD_{ha}^{-1}\E[\bg_{ha}(A)K_{ha}(A)\theta(A)]$ (which can be viewed as population version of local linear estimator $\be_1^{\top}\widehat{\bbeta}_h(a)$) with efficient influence function
\[
\begin{aligned}
     \phi_{ha}(\bZ)  
    = &\,\be_1^{\top}\bD_{ha}^{-1}\bg_{ha}(A)K_{ha}(A)\left(\varphi(\bZ)-\bg_{ha}^{\top}(A)\bD_{ha}^{-1}\E[\bg_{ha}(A)K_{ha}(A)\theta(A)]\right) \\
    &\,+ \be_1^{\top}\bD_{ha}^{-1}\int \bg_{ha}(t)K_{ha}(t)\tau(t,\bV)f(t) dt -\theta_h(a).
\end{aligned}
\]
Following \citet{kennedy2017non,takatsu2022debiased}, one can show the variance of $\widehat{\theta}(a)$ can be approximated by $\frac{1}{n} \Pb_n \left[\left(\widehat{\phi}_{ha}(\bZ)\right)^2\right]$ for
\[
\widehat{\phi}_{ha}(\bZ)=\be_1^{\top}\widehat{\bD}_{ha}^{-1}\bg_{ha}(A)K_{ha}(A)\left(\widehat{\varphi}(\bZ)-\bg_{ha}^{\top}(A)\widehat{\beta}_h(a)\right)+ \be_1^{\top}\widehat{\bD}_{ha}^{-1}\int \bg_{ha}(t)K_{ha}(t)\widehat{\tau}(t,\bV) d \Pb_n(t) -\widehat{\theta}(a).
\]

Finally, we compare the asymptotic variance in Theorem \ref{thm:surrogates-normality} with the asymptotic variance in \citet{kennedy2017non}, where the unlabeled dataset $\mathcal{U}$ and surrogates $\bS$ are unavailable. Consider the MCAR setting where $R$ is independent of all other variables so that $\rho(a,\bx) = \rho \in (0,1)$, and for simplicity assume $n_1 = n\rho$ to show how the surrogates and unlabeled data help to reduce the variance in this special setting. Note that since $R$ is independent of all other variables, we have $\operatorname{Var}(Y|A=a,\bX,R=1) = \operatorname{Var}(Y|A=a,\bX)$ and $ \mu(a,\bX,1) = \E[Y|A=a,\bX,R=1] = \E[Y|A=a,\bX] = \mu(a,\bX)$. The asymptotic variance of $\widehat{\theta}(a)$ in our setting (i.e. in Theorem \ref{thm:surrogates-normality}) is reduced to
\begin{equation}\label{eq:surrogates-var}
\frac{1}{nh} \E \left\{ \left[ \frac{\operatorname{Var}(Y|A=a,\bX)}{\rho}+\operatorname{Var}(\mu(a,\bX)|A=a, \bV) \right]w^2(a,\bV) \bigg | A=a  \right\}
\end{equation}
under the MCAR assumption (note the additional factor $\int K^2(u) du /f(a)$ is omitted since it appears in both settings). In the setting where the unlabeled data is unavailable, the asymptotic variance is shown to be
\begin{equation}\label{eq:supervised-var1}
    \frac{1}{n_1 h}\E \left\{ \operatorname{Var}(Y|A=a,\bV)w^2(a,\bV)|A=a \right \}
\end{equation}
in \citet{kennedy2017non}. By the property of conditional variance 
\[
\operatorname{Var}(Y|A=a,\bV) = \E[\operatorname{Var}(Y|A=a,\bX)|A=a, \bV]+\operatorname{Var}(\E[Y|A=a,\bX]|A=a,\bV),
\]
\eqref{eq:supervised-var1} can be re-written as
\begin{equation}\label{eq:supervised-var2}
\frac{1}{n_1 h}\E \left\{ (\operatorname{Var}(Y|A=a,\bX)+\operatorname{Var}(\mu(a,\bX)|A=a,\bV))w^2(a,\bV)|A=a \right \} .
\end{equation}
Comparing \eqref{eq:surrogates-var} with \eqref{eq:supervised-var2} we see the first term is the same since $n\rho = n_1$. However
the second term in \eqref{eq:supervised-var2} is improved by a factor of $\rho$ in \eqref{eq:surrogates-var}. This shows how the variance of the estimator is smaller after introducing unlabeled data and surrogate outcomes. The amount of improvement depends on the missing rate $1-\rho$ and $\operatorname{Var}(\mu(a,\bX)|A=a,\bV)$, which measures the variation of $\mu(A,\bV, \bS)$ that cannot be explained by $(A,\bV)$.

\section{Simulation Study}\label{sec:simulation}

\begin{figure*}[ht]
	\centering
	\subfigure[$n=500$]{
		\begin{minipage}[t]{0.45\linewidth}
			\centering
			\includegraphics[width=3in]{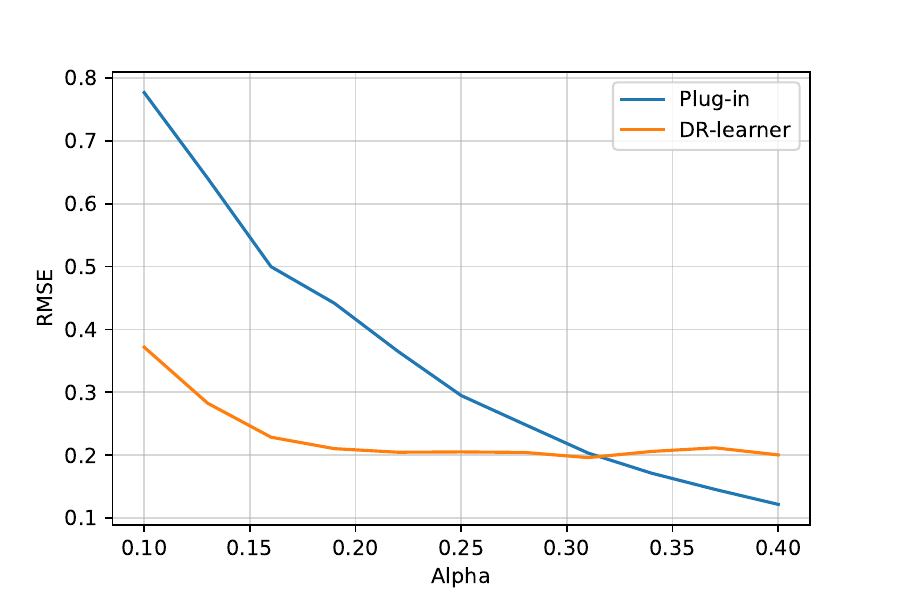}
	\end{minipage}}
	\subfigure[$n=2000$]{
		\begin{minipage}[t]{0.45\linewidth}
			\centering
			\includegraphics[width=3in]{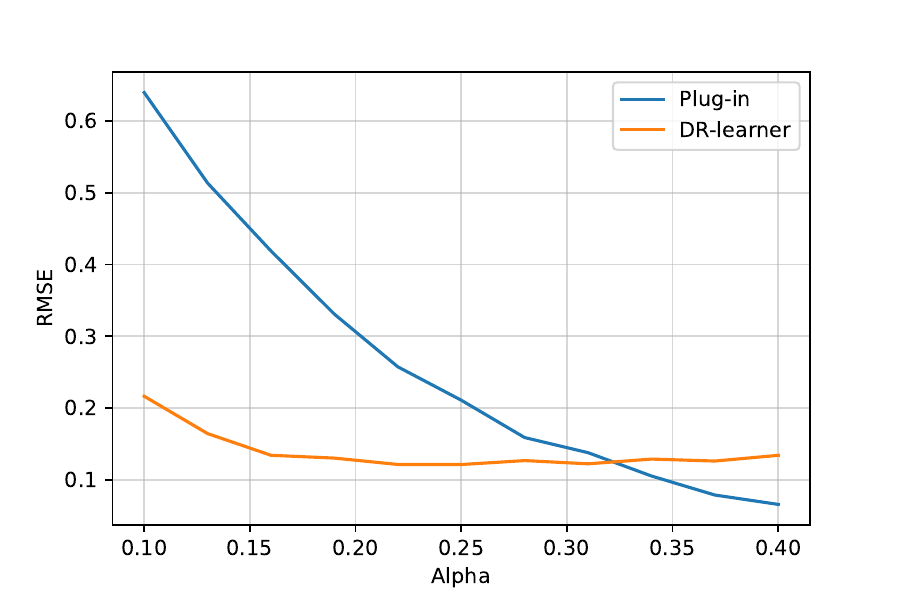}
	\end{minipage}}
	\centering
	\caption{Root mean square error Versus $\alpha$, where $n^{-\alpha}$ is the estimation error of the nuisance functions.}
	\label{fig:rmse_alpha}
\end{figure*}

In this section we use simulations to evaluate the performance of the proposed methods. We will illustrate the advantage of doubly robust estimation over naive plug-in style estimators. Consider the following data-generating process: The covariates $\bV$ have a multi-variate Gaussian distribution
\[
\bV = (V_1, V_2, V_3, V_4)^{\top} \sim N(\mathbf{0}, \mathbf{I}_4) .
\]
Conditioning on $\bV$, the continuous treatment $A$ has normal distribution $N(\lambda(\bV),1)$ with
\[
\lambda(\bV) = 1+0.2V_1+0.2V_2 - 0.2V_3 + 0.3V_4.
\]
The surrogates $\bS$ has a normal distribution
\[
\bS = (S_1,S_2)^{\top} \sim N(\mathbf{0}, \mathbf{I}_2). 
\]
The indicator of whether the outcome is observed or not $R$ is Bernoulli$(0.5)$ (so we assume a missing completely at random mechanism and $\rho = 0.5$). Finally the outcome $Y$ conditioning on $A,\bX,R=1$ has a normal distribution $N(\mu(A,\bX,1), 1)$ with
\[
\mu(A,\bX,1) =1 + (0.1,-0.1)^{\top} \bS + (0.2,0.2,0.3,-0.1)^{\top}\bV+ A(1-0.1V_1 + 0.1V_3) - A^2.
\]
By direct calculations we have
\[
\tau(A,\bV) =1+ (0.2,0.2,0.3,-0.1)^{\top}\bV+ A(1-0.1V_1 + 0.3V_3) - A^2,
\]
The dose-response function of interest is
\[
\theta(A) = 1 + A-A^2. 
\]
To illustrate the performance of two estimators with different nuisance estimation errors we will manually set the estimation error, which is applicable for simulation purposes. For a fixed $\alpha$ we let $\epsilon_1, \dots, \epsilon_4 \sim N(n^{-\alpha}, n^{-2\alpha})$ and set $\widehat{\lambda}(\bV)= \lambda(\bV) + \epsilon_1$, the estimated conditional density of $A$ is $ N\left(\widehat{\lambda}(\bV),1 \right), \widehat{\mu}(A,\bX,1) = \mu(A,\bX,1) + \epsilon_2, \widehat{\tau}(A,\bV) = \tau(A,\bV) + \epsilon_3, \text{logit} \left(\widehat{\rho} \right) = \text{logit}(\rho) + \epsilon_4$. Such estimates guarantee the nuisance estimation error is of order $n^{-\alpha}$. After we generate a sample of size $n$, the plug-in style estimator is defined as $\widehat{\theta}(a) = \frac{1}{n} \sum_{i=1}^n \widehat{\tau}(a,\bV_i)$. To implement the doubly robust estimator we split the sample into two parts $D,T$ (since the nuisance estimators are given there is no need to split the sample into three parts as in Algorithm \ref{alg:DR-surrogates}). We use $D$ to construct estimator of the initial estimator $\widehat{\theta}_0(a) = \frac{1}{|D|}\sum_{i \in D}\widehat{\tau}(a,\bV_i)$, marginal density $\widehat{f}(a) = \frac{1}{|D|}\sum_{i \in D}\widehat{\pi}(a|\bV_i)$ and select the optimal bandwidth $h^*$ (i.e. construct pseudo-outcomes and run LOOCV as in equation \eqref{eq:loocv} on $D$). We then construct pseudo-outcomes on $T$ and perform local linear estimation using the optimal bandwidth $h^*$. Finally, the roles of $D$ and $T$ are exchanged to obtain another estimator and we average the two estimates as the final doubly robust estimator. For sample size $n \in \{500, 2000\}$ and convergence rate $\alpha \in \{ 0.1, 0.13, \dots, 0.4\}$, we repeat the data generation and estimation process $M=500$ times. We will aim at estimating $\theta(1)$ and compare the RMSE$=\left \{\frac{1}{M} \sum_{m=1}^M \left[\widehat{\theta}_m(1)-\theta(1)\right]^2 \right \}^{1/2}$ of plug-in estimator and doubly robust estimator, where $\widehat{\theta}_m(1)$ is the estimate from $m$-th repetition. The results are summarized in Figure \ref{fig:rmse_alpha}.


As we see in Figure \ref{fig:rmse_alpha}, if the nuisance estimation error is large ($\alpha$ is small), the doubly robust estimator outperforms the naive plug-in estimator. This can be explained by the second-order bias term of the doubly robust estimator in Proposition \ref{prop:surrogates-bias}, i.e., the conditional bias is the product of nuisance estimation errors and is ``doubly small". On the other hand, the plug-in style estimator inherits the slow convergence rate of $\widehat{\tau}$. As $\alpha$ increases, the estimators of nuisance functions are more accurate, and plug-in style estimators finally outperform the doubly robust estimator because the doubly robust estimator may suffer from accumulating error in constructing pseudo-outcomes, bandwidth selection, and local linear regression, which dominates the conditional bias when nuisance estimation is accurate enough. In real applications, there are typically many covariates and parametric models for nuisance functions may not be correct. The convergence rate of nonparametric nuisance estimation can be slow when the dimension of covariates is large so the doubly robust estimator with a smaller bias is recommended for use. Additional simulation results and a real data example are provided in the supplementary materials. 

\section{Discussion}\label{sec:discussion}

In this work, we study the estimation of continuous treatment effects when there is limited access to the primary outcome of interest but auxiliary information on surrogate outcomes is available. We propose a doubly robust estimator that is less sensitive to nuisance estimation error and hence incorporates flexible nonparametric machine learning methods. Although nonparametric machine learning methods usually suffer from slow convergence rates, they are widely used in nuisance function estimation, especially when practitioners do not have sufficient domain knowledge to justify parametric models. Our doubly robust estimator facilitates the application of nonparametric methods and enjoys optimal estimation rates under mild conditions. Asymptotic normality is further established, which enables researchers to construct confidence intervals and perform statistical inference. We also show how incorporating information on surrogate outcomes improves the variance, compared with methods solely based on labeled data, as in \citet{kennedy2017non}. In summary, our methodology provides a robust and efficient approach to leverage surrogate outcomes in continuous treatment effect estimation. However, our method could not deal with the case where only covariate is available in the unlabeled dataset and continuous treatment is also missing (corresponds to the generalizability problem as in \citet{dahabreh2019generalizing}). It is also interesting to extend our results to studies with multiple outcomes \citep{kennedy2019estimating, du2024causal}. We leave these problems for future investigation.



\bibliographystyle{apalike}
\bibliography{references}

\newpage

\appendix

\section{Background on Efficiency Theory}

As discussed in Section \ref{sec:identification}, the plug-in estimator suffers from first-order error and is sensitive to the estimation accuracy of nuisance functions. To address this problem, one can derive the efficient influence function of the target functional ($\E[\theta(A)]$ in our problem), based on which a one-step estimator can be obtained to reduce the bias. The efficient influence function is critical in non-parametric efficiency theory \citep{bickel1993efficient, tsiatis2006semiparametric, van2000asymptotic,laan2003unified, kennedy2022semiparametric}. Mathematically, the influence function is the derivative of the target statistical functional in a
Von Mises expansion (i.e., distributional Taylor expansion). In the discrete case, it coincides with the Gateaux derivative of the functional when the contamination distribution is a point mass. Influence functions are important in different respects. First, the variance of the influence function is equal to the efficiency bound of the target statistical functional, which characterizes the inherent estimation difficulty of the target functional and provides a benchmark to compare against when we construct estimators. Moreover, it allows us to correct for first-order bias in the plug-in estimator and obtain doubly robust-style estimators, which enjoy appealing statistical properties even if non-parametric methods with relatively slow rates are used in nuisance estimation.

Suppose the statistical functional of interest $\psi=\psi(\Pb)$ admits the first-order Von Mises expansion. Mathematically, we have
\begin{equation}\label{eq:von-mises}
    \psi(\hat{\mathbb{P}})-\psi(\mathbb{P})=-\int \phi_1(\bz,\hat{\mathbb{P}}) d \mathbb{P}(\bz)+R_2(\hat{\mathbb{P}}, \mathbb{P}),
\end{equation}
where $\phi_1(\bz,\Pb)$ is the influence function of $\psi(\Pb)$, $\psi(\hat{\mathbb{P}})$ is the plug-in estimator and $R_2(\hat{\mathbb{P}}, \mathbb{P})$ is the second-order reminder. Under regularity conditions, Von Mises expansion implies the pathwise differentiability 
\[
\left.\frac{\partial}{\partial \epsilon} \psi\left(\mathbb{P}_\epsilon\right)\right|_{\epsilon=0}=\int \phi_1(\bz ; \mathbb{P}) s_\epsilon(\bz) d \mathbb{P}(z)
\]
where $s_\epsilon(\bz)=\left.\frac{\partial}{\partial \epsilon} \log p_\epsilon(\bz)\right|_{\epsilon=0}$ is
the submodel score function. \eqref{eq:von-mises} suggests that the plug-in estimator has first-order bias $-\int \phi_1(\bz,\hat{\mathbb{P}}) d \mathbb{P}(\bz)$. Equivalently, we can write
\[
\psi(\mathbb{P})=\psi(\hat{\mathbb{P}})+\int \phi_1(\bz,\hat{\mathbb{P}}) d \mathbb{P}(\bz) +R_2(\hat{\mathbb{P}}, \mathbb{P}),
\]
which motivates us to correct for the first-order bias and arrive at the doubly robust estimator
\[
\hat{\psi}^{dr} = \psi(\hat{\mathbb{P}})+\mathbb{P}_n\{\phi_1(\bZ,\hat{\mathbb{P}})\}.
\]
Under further empirical process assumptions or sample splitting assumptions, we can show the dominating term in the conditional bias of $\hat{\psi}^{dr}$ is $R_2(\hat{\mathbb{P}}, \mathbb{P})$, which is usually a second-order error term and depends on the product of convergence rates of nuisance functions. We refer the readers to \citet{kennedy2016doublerobust, kennedy2022semiparametric} for a more complete review. In our problem, the estimand $\theta(a)$ in \eqref{eq:identification} is not pathwise differentiable and the ideas above do not apply directly. However, the pseudo-outcome $\varphi(\bZ)$ is obtained by deriving the influence function of $\E[\theta(A)]$. The readers are referred to Section 3.1 of \citet{kennedy2017non} for more discussion on the connection between the pseudo-outcome of $\theta(a)$ and the influence function of $\E[\theta(A)]$.

\section{Proofs}
\subsection{Proof of Theorem \ref{thm:identification}}
\begin{proof}
     \[
    \begin{aligned}
        &\, \E[Y^a] \\
        = &\, \E[\E(Y^a|\bV, \bS^a] \\
        = &\, \E[\E(Y^a|A=a, \bV, \bS^a)] \\ 
        = &\, \E[\E(Y^a|A=a, \bV, \bS^a, R=1)] \\
        = &\, \E\{\E[\E(Y^a|A=a, \bV, \bS^a, R=1)|\bV]\}\\
        = &\, \E\{\E[\E(Y^a|A=a, \bV, \bS^a, R=1)|A=a, \bV]\}\\
        = & \, \E\{\E[\E(Y|A=a, \bV, \bS, R=1)|A=a, \bV]\},
    \end{aligned}
    \]
    where the first and fourth equations follow from the property of conditional expectation. The second equation holds since Assumption \ref{asm:surrogates-A} implies $Y^a \independent A | \bV, \bS^a$. The third equation follows from Assumption \ref{asm:surrogates-R}. The fifth equation follows from Assumption \ref{asm:surrogates-A} (specifically $\bS^a \independent A | \bV$). The last equation follows from Assumption \ref{asm:consistency}. Note that positivity is implicitly assumed to guarantee the conditional expectations are well-defined.
\end{proof}

\subsection{Proof of Proposition \ref{prop:surrogates-dr}}
\begin{proof}
    If $\bar{\mu} = \mu$ we have
    \[
    \begin{aligned}
        \varphi(\bZ; \mu, \bar{\pi}, \bar{\rho}) =&\, \left[\frac{R(Y-{\mu}(A,\bX,1))}{\bar{\rho}(A,\bX)} + {\mu}(A, \bX, 1) - \E[{\mu}(A,\bX,  1)|A,\bV] \right]\frac{\int_{\mathcal{V}} \bar{\pi}(A|\bv) d \Pb(\bv)}{\bar{\pi}(A|\bV)} \\
    &\, + \theta(A).
    \end{aligned}
    \]
    Note that 
    \[
    \E\left[ \frac{R(Y-{\mu}(A,\bX,1))}{\bar{\rho}(A,\bX)}  \bigg | A=a ,\bX, R \right] = \frac{R}{\bar{\rho}(a,\bX)} \E[Y-\mu(a,\bX,  1) | A=a,\bX, R=1] = 0,
    \]
    \[
    \E[{\mu}( A,\bX, 1) - \E[{\mu}(A,\bX,  1)|A, \bV]|A=a, \bV]=0.
    \]
    These two equations imply 
    \[
    \E\left \{  \left[\frac{R(Y-{\mu}(A,\bX,1))}{\bar{\rho}(A,\bX)} + {\mu}(A,\bX,  1) - \E[{\mu}(A,\bX,  1)|A,\bV] \right]\frac{\int_{\mathcal{V}} \bar{\pi}(A|\bv) d \Pb(\bv)}{\bar{\pi}(A|\bV)} \bigg | A=a\right\} = 0
    \]
    and hence
    \[
    \E[\varphi(\bZ; \mu, \bar{\pi}, \bar{\rho})|A=a] = \theta(a).
    \]
    If $(\bar{\pi}, \bar{\rho}) = (\pi, \rho)$, 
    \[
    \begin{aligned}
    \varphi(\bZ; \bar{\mu}, {\pi}, {\rho}) =&\, \left[\frac{R(Y-\bar{\mu}(A,\bX,1))}{{\rho}(A,\bX)} + \bar{\mu}(A,\bX,  1) - \E[\bar{\mu}(A,\bX,  1)|A,\bV] \right]\frac{\int_{\mathcal{V}} {\pi}(A|\bv) d \Pb(\bv)}{{\pi}(A|\bV)} \\
    &\, + \int_{\mathcal{V}} \E[\bar{\mu}(A,\bX,1)|A,\bV=\bv] d \Pb(\bv).
\end{aligned}
    \]
    Note that
    \[
    \E\left[ \frac{R(Y-\bar{\mu}(A,\bX,1))}{\rho(A,\bX)}  \bigg |A=a, \bX, R \right] = \frac{R(\mu(a,\bX,1) - \bar{\mu}(a,\bX,1))}{\rho(a,\bX)},
    \]
    \[
    \E\left[ \frac{R(Y-\bar{\mu}(A,\bX,1))}{\rho(A,\bX)}  \bigg |A=a, \bX \right] = \E \left[ \frac{R(\mu(a,\bX,1) - \bar{\mu}(a,\bX,1))}{\rho(a,\bX)} \bigg | A=a, \bX\right] = \mu(a,\bX,1) - \bar{\mu}(a,\bX,1).
    \]
    By the property of conditional expectation, we have
    \[
    \begin{aligned}
        &\, \E \left\{\left[\frac{R(Y-\bar{\mu}(A,\bX,1))}{{\rho}(A,\bX)} + \bar{\mu}(A,\bX,  1) - \E[\bar{\mu}( A,\bX, 1)|A,\bV] \right]\frac{\int_{\mathcal{V}} {\pi}(A|\bv) d \Pb(\bv)}{{\pi}(A|\bV)} \bigg| A=a, \bV\right\} \\
        = &\, \E \left\{ \mu(a,\bX,1)-\E[\bar{\mu}( a, \bX,1)|A=a,\bV]  \big|A=a, \bV  \right\} \frac{\int_{\mathcal{V}} {\pi}(a|\bv) d \Pb(\bv)}{{\pi}(a|\bV)} \\
        = &\, \E [ \mu(a,\bX,1)-\bar{\mu}(a,\bX,  1)\big| A=a,\bV ] \frac{\int_{\mathcal{V}} {\pi}(a|\bv) d \Pb(\bv)}{{\pi}(a|\bV)}.
    \end{aligned}
    \]
    Due to the following equation on the measure:
    \begin{equation}\label{eq:measure-relation}
        d \Pb (\bv|a) = \frac{\pi(a|\bv) d \Pb(\bv)}{\int_{\mathcal{V}} {\pi}(a|\bv) d \Pb(\bv)}, 
    \end{equation}
    we can write
    \[
    \begin{aligned}
        &\, \E \left\{\left[\frac{R(Y-\bar{\mu}(A,\bX,1))}{{\rho}(A,\bX)} + \bar{\mu}(A,\bX,  1) - \E[\bar{\mu}(A,\bX,  1)|A,\bV] \right]\frac{\int_{\mathcal{V}} {\pi}(A|\bv) d \Pb(\bv)}{{\pi}(A|\bV)} \bigg| A=a \right\} \\
        = &\, \int_{\mathcal{V}} \E [ \mu(a,\bX,1)-\bar{\mu}(a,\bX,  1)\big| A=a,\bV=\bv  ] \frac{\int_{\mathcal{V}} {\pi}(a|\bv) d \Pb(\bv)}{{\pi}(a|\bv)} d \Pb (\bv|a) \\
        = &\, \int_{\mathcal{V}} \E [ \mu(a,\bX,1)-\bar{\mu}(a,\bX,  1)\big| A=a,\bV=\bv]  d \Pb (\bv) .
    \end{aligned}
    \]
    Finally we get
    \[
    \begin{aligned}
        &\, \E[\varphi(\bZ; \bar{\mu}, {\pi}, {\rho})|A=a] \\
        = & \, \int_{\mathcal{V}} \E [ \mu(a,\bX,1)-\bar{\mu}(a,\bX,  1)\big|A=a, \bV=\bv ]  d \Pb (\bv) + \int_{\mathcal{V}} \E[\bar{\mu}(a,\bX,1)|A=a,\bV=\bv ] d \Pb(\bv) \\
        = &\, \int_{\mathcal{V}} \E[{\mu}(a,\bX,1)|A=a,\bV=\bv] d \Pb(\bv) \\
        = &\, \theta(a).
    \end{aligned}
    \]
\end{proof}

\subsection{Proof of Theorem \ref{thm:DR-surrogates}}
\begin{proof}
    By Theorem 1 in \citet{kennedy2023towards}, the linear smoother is stable if the variance $\operatorname{Var}(\varphi(\bZ)|A=a)$ is bounded away from $0$, i.e.
    \[
    \widehat{\theta}(a) - \widetilde{\theta}(a)- \widehat{\E}_n[\widehat{b}(A)|A=a] = o_{\Pb}(R_n(a))
    \]
    if $\sup_{\bz}|\widehat{\varphi}(\bz) - \varphi(\bz)| = o_{\Pb}(1)$.
\end{proof}

\subsection{Proof of Proposition \ref{prop:surrogates-bias}}
\begin{proof}
    For abbreviations, we will omit conditioning on $D^n$ in our notation but all the expectations in this part are conditioning on $D^n$ (recall such expectation is denoted using $\Pb$).
    Note that 
    \[
    \E[\varphi(\bZ)|A=a] = \theta(a)
    \]
    by Proposition \ref{prop:surrogates-dr}. By the property of conditional expectation
    \begin{equation}\label{eq:bias1}
        \begin{aligned}
        &\, \Pb \left \{ \left[ \frac{R(Y-\widehat{\mu}(a,\bX,1))}{\widehat{\rho}(a,\bX)} + \widehat{\mu}(a,\bX,1) -\widehat{\tau}(a,\bV) \right] \widehat{w}(a,\bV)\bigg | A=a \right\} \\
        = &\, \Pb \left \{ \left[ \frac{R(\mu(a,\bX,1)-\widehat{\mu}(a,\bX,1))}{\widehat{\rho}(a,\bX)} + \widehat{\mu}(a,\bX,1) -\widehat{\tau}(a,\bV) \right] \widehat{w}(a,\bV)\bigg | A=a \right\} \\
        = & \, \Pb \left \{ \left[ \frac{\rho(a,\bX)(\mu(a,\bX,1)-\widehat{\mu}(a,\bX,1))}{\widehat{\rho}(a,\bX)} + \widehat{\mu}(a,\bX,1) -\widehat{\tau}(a,\bV) \right] \widehat{w}(a,\bV)\bigg | A=a \right\} \\
        = & \, \Pb \left \{ \left[ \left( 1-\frac{\rho(a,\bX)}{\widehat{\rho}(a,\bX)} \right)(\widehat{\mu}(a,\bX,1)-\mu(a,\bX,1)) + {\mu}(a,\bX,1) -\widehat{\tau}(a,\bV) \right] \widehat{w}(a,\bV)\bigg | A=a \right\} \\
        = &\, \Pb \left \{ \left[ \left( 1-\frac{\rho(a,\bX)}{\widehat{\rho}(a,\bX)} \right)(\widehat{\mu}(a,\bX,1)-\mu(a,\bX,1))  \right] \widehat{w}(a,\bV)\bigg | A=a \right\} + \Pb [(\tau(a,\bV) - \widehat{\tau}(a,\bV))\widehat{w}(a,\bV)|A=a]
    \end{aligned}
    \end{equation}
    where the first equation follows by conditioning on $\bX,R,A=a$, the second equation follows from conditioning on $\bX,A=a$ and the last equation follows from conditioning on $\bV,A=a$. Further note that $\theta(a) = \E[\tau(a,\bV)]$ and
    \begin{equation}\label{eq:bias2}
        \Pb\left[\widehat{\theta}_0(a) \right]-\theta(a) = \frac{1}{n}\sum_{i \in D_2^n}\widehat{\tau}(a,\bV_i) - \Pb[\widehat{\tau}(a,\bV)] + \Pb[\widehat{\tau}(a,\bV) - \tau(a,\bV)].
    \end{equation}
    By equation \ref{eq:measure-relation}
    \[
    \Pb[\widehat{\tau}(a,\bV) - \tau(a,\bV)] = \int_{\mathcal{V}} (\widehat{\tau}(a,\bv) - \tau(a,\bv)) d \Pb(\bv) = \int_{\mathcal{V}} (\widehat{\tau}(a,\bv) - \tau(a,\bv)) w(a,\bv)d \Pb(\bv|A=a).
    \]
    Add equation \ref{eq:bias1} and equation \ref{eq:bias2} together we have
    \[
    \begin{aligned}
        \widehat{b}(a) =&\, \Pb \left \{ \left[ \left( 1-\frac{\rho(a,\bX)}{\widehat{\rho}(a,\bX)} \right)(\widehat{\mu}(a,\bX,1)-\mu(a,\bX,1))  \right] \widehat{w}(a,\bV)\bigg | A=a \right\}\\
        &\, + \Pb [(\tau(a,\bV) - \widehat{\tau}(a,\bV))(\widehat{w}(a,\bV)-w(a,\bV))|A=a] + \frac{1}{n}\sum_{i \in D_2^n}\widehat{\tau}(a,\bV_i) - \Pb[\widehat{\tau}(a,\bV)].
    \end{aligned}
    \]
    The bound on $\widehat{b}$ then follows from Cauchy-Schwarz's inequality. 

    For linear smoother $\widehat{\E}_n$, we have
    \[
    \left |\widehat{\E}_n \left[\widehat{b}(A)|A=a\right] \right| = \left | \sum_{i \in T^n} W_i(a;\bA^n) \widehat{b}(A_i) \right | \leq  \sup_{t \in N(a)} |\widehat{b}(t) | \sum_{i \in T^n} |W_i(a;\bA^n)| \leq C \sup_{t \in N(a)} |\widehat{b}(t) |
    \]
\end{proof}

\subsection{Proof of Theorem \ref{thm:surrogates-normality}}
\begin{proof}
    We will prove the results with the following decomposition.
     Let $\widetilde{\theta}(a) = \be_1^{\top} \widehat{\bD}_{ha}^{-1}\Pb_n[\bg_{ha}(A)K_{ha}(A)\varphi(\bZ)]$ be the oracle estimator. We write
    \[
    \begin{aligned}
        \widehat{\theta}(a) - \theta(a) = &\, \widetilde{\theta}(a) - \theta(a) + \be_1^{\top} \widehat{\bD}_{ha}^{-1}\Pb_n[\bg_{ha}(A)K_{ha}(A)(\widehat{\varphi}(\bZ) - \varphi(\bZ))]\\
        =: &\, \widetilde{\theta}(a) - \theta(a) + R_1 + R_2
    \end{aligned}
    \]
    where $R_1 = \be_1^{\top}\widehat{\bD}_{ha}^{-1} (\Pb_n - \Pb)[\bg_{ha}(A)K_{ha}(A)(\widehat{\varphi}(\bZ) -\varphi(\bZ))]$ and $R_2 = \be_1^{\top}\widehat{\bD}_{ha}^{-1} \Pb[\bg_{ha}(A)K_{ha}(A)(\widehat{\varphi}(\bZ) -\varphi(\bZ))]$. 
    
    \textbf{Step 1: The CLT term.} The existing results on local linear estimator \citep{fan1994robust, fan1996local} imply 
    \[
    \sqrt{nh} \left( \widetilde{\theta}(a) - \theta(a)-\frac{h^2 \theta''(a) \int u^2 K(u)du}{2} \right) \overset{d}{\rightarrow} N \left( 0, \frac{\sigma^2(a) \int K^2(u) du}{f(a)} \right)
    \]
    under the conditions stated in the theorem. The conditional variance can be computed as follows:
    \[
    \begin{aligned}
        \sigma^2(a) = &\, \operatorname{Var}(\varphi(\bZ)|A=a) \\
        = &\, \E[(\varphi(\bZ)-\theta(a))^2 |A=a] \\
        =&\, \E \left\{ \left[ \frac{R(Y-\mu(a,\bX,1))}{\rho(a,\bX)} + \mu(a,\bX,1) - \tau(a,\bV) \right]^2 w^2(a,\bV) \bigg | A=a \right\} \\
        = &\, \E \left \{ \left[ \frac{R(Y-\mu(a,\bX,1))^2}{\rho^2(a,\bX)} +(\mu(a,\bX,1)-\tau(a,\bV))^2 \right]w^2(a,\bV)\bigg |A=a \right\},
    \end{aligned}
    \]
    where the last equation follows since by conditioning on $A=a, \bX, R$ one can show
    \[
    \E \left \{ \left[ \frac{R(Y-\mu(a,\bX,1))(\mu(a,\bX,1)-\tau(a,\bV))}{\rho(a,\bX)}\right]w^2(a,\bV)\bigg |A=a \right\}=0.
    \]
    For the first term in $\sigma^2(a)$
    \[
    \begin{aligned}
        &\,\E \left \{ \frac{R(Y-\mu(a,\bX,1))^2w^2(a,\bV)}{\rho^2(a,\bX)} \bigg| A=a \right\} \\
        = &\, \E \left \{ \frac{R\operatorname{Var}(Y|A=a,\bX,R=1) w^2(a,\bV)}{\rho^2(a,\bX)} \bigg | A=a \right\} \\
        = &\, \E \left \{ \frac{\operatorname{Var}(Y|A=a,\bX,R=1)w^2(a,\bV)}{\rho(a,\bX)} \bigg | A=a \right\},
    \end{aligned}
    \]
    where the first equation follows from conditioning on $A=a,\bX, R$ and the second equation follows from conditioning on $A=a,\bX$. For the second term in $\sigma^2(a)$ we have
    \[
    \E \left[ (\mu(a,\bX,1)-\tau(a,\bV))^2 w^2(a,\bV)|A=a \right] = \E \left[ \operatorname{Var}(\mu(a,\bX,1)|A=a,\bV)w^2(a,\bV) |A=a\right].
    \]
    Hence 
    \[
    \sigma^2(a) = \E \left \{ \left[\frac{\operatorname{Var}(Y|A=a,\bX,R=1)}{\rho(a,\bX)} +\operatorname{Var}(\mu(a,\bX,1)|A=a,\bV) \right]w^2(a,\bV) \bigg | A=a \right\}
    \]
    \textbf{Step 2: Bounding $R_1$. }Then we proceed to analyze $R_1 = \be_1^{\top}\widehat{\bD}_{ha}^{-1} (\Pb_n - \Pb)[\bg_{ha}(A)K_{ha}(A)(\widehat{\varphi}(\bZ) -\varphi(\bZ))]$. We first show $\be_1^{\top}\widehat{\bD}_{ha}^{-1} = O_{\Pb}(1)$. Recall $\widehat{\bD}_{ha}= \Pb_n [\bg_{ha}(A)K_{ha}(A)\bg_{ha}^{\top}(A)] \in \mathbb{R}^{2\times 2}$. Note that $\Pb_n[K_{ha}(A)]$ is the kernel density estimator of $f(a)$ and standard results in the literature show
    \[
    \E[(\Pb_n[K_{ha}(A)] - f(a))^2] = O\left(h^2 + \frac{1}{nh}\right) = o(1),
    \]
    which implies $\widehat{D}_{ha,11} =  \Pb_n[K_{ha}(A)] \overset{P}{\rightarrow} f(a)$. For the element of $\widehat{\bD}_{ha}$ not on the diagonal
    \[
    \E\left[\left( \frac{A-a}{h} \right)K_{ha}(A)\right] = \int \frac{t-a}{h} \frac{1}{h}K\left(\frac{t-a}{h}\right) f(t) dt = \int u K(u)f(a+hu) du.
    \]
    So we have
    \[
    \left|\E\left[\left( \frac{A-a}{h} \right)K_{ha}(A)\right]- \int u K(u)f(a) du \right|  \leq \int |u|K(u)|f(a+hu)-f(a)| \lesssim h \int u^2 K(u) du \rightarrow 0. 
    \]
    Since $K$ is symmetric around $0$ we have $\int u K(u) du = 0$ and hence 
    \[
    \E\left[\left( \frac{A-a}{h} \right)K_{ha}(A)\right] = O(h). 
    \]
    Further notice
    \[
    \begin{aligned}
        &\, \operatorname{Var} \left(\Pb_n \left[ \left(\frac{A-a}{h}\right) K_{ha}(A) \right] \right)\\
        =&\, \frac{1}{n} \operatorname{Var} \left(  \left(\frac{A-a}{h}\right) K_{ha}(A) \right)\\
        \leq &\, \frac{1}{n} \E \left[ \left(\frac{A-a}{h}\right)^2 K_{ha}^2(A) \right] \\
        = & \, \frac{1}{n} \int \left(\frac{t-a}{h}\right)^2 \frac{1}{h^2}K^2\left(\frac{t-a}{h} \right) f(t) dt \\
        = & \, \frac{1}{nh} \int u^2 K^2(u)f(a+hu)du \\
        \leq &\, \frac{\|f\|_{\infty}}{nh} \int u^2 K^2(u)du = O\left( \frac{1}{nh} \right).
    \end{aligned}
    \]
    Hence we have
    \[
    \E  \left \{ \left\{ \Pb_n \left[ \left(\frac{A-a}{h}\right) K_{ha}(A) \right] \right\}^2 \right \} = O\left(h^2 + \frac{1}{nh}\right) = o(1),
    \]
    which implies $\widehat{D}_{ha,12} =\Pb_n \left[ \left(\frac{A-a}{h}\right) K_{ha}(A) \right] \overset{P}{\rightarrow} 0$. Finally
    \[
    \E\left[\left( \frac{A-a}{h} \right)^2K_{ha}(A)\right] = \int \left(\frac{t-a}{h} \right)^2\frac{1}{h}K\left(\frac{t-a}{h}\right) f(t) dt = \int u^2 K(u)f(a+hu) du.
    \]
    So we have
    \[
    \left|\E\left[\left( \frac{A-a}{h} \right)^2 K_{ha}(A)\right]- \int u^2 K(u)f(a) du \right|  \leq \int u^2K(u)|f(a+hu)-f(a)| \lesssim h \int |u|^3 K(u) du \rightarrow 0. 
    \]
    One can similarly show 
    \[
    \operatorname{Var}\left(\left( \frac{A-a}{h} \right)^2 K_{ha}(A) \right) = O(1/h)
    \]
    and hence
    \[
    \E  \left \{ \left\{ \Pb_n \left[ \left(\frac{A-a}{h}\right)^2 K_{ha}(A) \right] - \int u^2 K(u)f(a) du \right\}^2 \right \}  = O\left(h^2 + \frac{1}{nh}\right) = o(1), 
    \]
    which implies $\widehat{D}_{ha,22} =\Pb_n \left[ \left(\frac{A-a}{h}\right)^2 K_{ha}(A) \right] \overset{P}{\rightarrow} f(a) \int u^2K(u) du$. Hence we have
    \[
    \widehat{\bD}_{ha}^{-1} \overset{P}{\rightarrow} \text{diag} \left \{ f(a)^{-1}, f(a)^{-1} \left(\int u^2K(u) du\right)^{-1}\right\}. 
    \]
    This implies $\be_1^{\top}\widehat{\bD}_{ha}^{-1} = O_{\Pb}(1)$. Then we consider $(\Pb_n - \Pb)[\bg_{ha}(A)K_{ha}(A)(\widehat{\varphi}(\bZ) -\varphi(\bZ))]$. By Lemma 2 in \citet{kennedy2020sharp} we have for $j=1,2$
    \[
    (\Pb_n - \Pb)[g_{ha,j}(A)K_{ha}(A)(\widehat{\varphi}(\bZ) -\varphi(\bZ))] = O_{\Pb}\left( \frac{\|g_{ha,j}(A) K_{ha}(A)(\widehat{\varphi}(\bZ) - \varphi(\bZ))\|_2}{\sqrt{n}} \right)
    \]
    Note that
    \[
    \begin{aligned}
        &\,\|g_{ha,j}(A) K_{ha}(A)(\widehat{\varphi}(\bZ) - \varphi(\bZ))\|_2^2 \\
        =&\, \Pb\left[g_{ha,j}^2(A) K_{ha}^2(A)(\widehat{\varphi}(\bZ) - \varphi(\bZ))^2 \right] \\
        \leq &\, \|\widehat{\varphi}-\varphi\|_{\infty}^2 \int \left( \frac{t-a}{h}\right)^{2(j-1)} \frac{1}{h^2} K^2 \left( \frac{t-a}{h}\right) f(t) dt \\
        \leq &\, \frac{\|\widehat{\varphi}-\varphi\|_{\infty}^2}{h} \int u^{2(j-1)}K^2(u)f(a+hu) du \\
        \lesssim &\, \frac{\|\widehat{\varphi}-\varphi\|_{\infty}^2}{h}.
    \end{aligned}
    \]
    This together with $\|\widehat{\varphi}-\varphi\|_{\infty}= o_{\Pb}(1)$ implies 
    \[
     (\Pb_n - \Pb)[g_{ha,j}(A)K_{ha}(A)(\widehat{\varphi}(\bZ) -\varphi(\bZ))] = O_{\Pb}\left( \frac{\|\widehat{\varphi}-\varphi\|_{\infty}}{\sqrt{nh}} \right) = o_{\Pb} \left( \frac{1}{\sqrt{nh}} \right).
    \]
    We conclude that $R_1 = o_{\Pb} \left( \frac{1}{\sqrt{nh}} \right)$.
    
    \textbf{Step 3: Bounding $R_2$. } The last step is to bound $R_2 = \be_1^{\top}\widehat{\bD}_{ha}^{-1} \Pb[\bg_{ha}(A)K_{ha}(A)(\widehat{\varphi}(\bZ) -\varphi(\bZ))]$. Since $\be_1^{\top}\widehat{\bD}_{ha}^{-1} = O_{\Pb}(1)$ we only need to consider
    \[
    \Pb[\bg_{ha}(A)K_{ha}(A)(\widehat{\varphi}(\bZ) -\varphi(\bZ))] = \int \bg_{ha}(t)K_{ha}(t) \Pb[\widehat{\varphi}(\bZ) - \varphi(\bZ)|A=t] f(t) dt
    \]
    
    In Proposition \ref{prop:surrogates-bias} we show $\widehat{b}(a)$ is equal to
    \[
    \begin{aligned}
         \Pb[\widehat{\varphi}(\bZ) - \varphi(\bZ)|A=t]=&\, \Pb \left \{ \left[ \left( 1-\frac{\rho(t,\bX)}{\widehat{\rho}(t,\bX)} \right)(\widehat{\mu}(t,\bX,1)-\mu(t,\bX,1))  \right] \widehat{w}(t,\bV)\bigg | A=t \right\}\\
        &\, + \Pb [(\tau(t,\bV) - \widehat{\tau}(t,\bV))(\widehat{w}(t,\bV)-w(t,\bV))|A=t] + \frac{1}{n}\sum_{i \in D_2^n}\widehat{\tau}(t,\bV_i) - \Pb[\widehat{\tau}(t,\bV)]
    \end{aligned}
    \]
    Plug into the equation above we have
    \[
    \begin{aligned}
        &\, \int g_{ha,j}(t)K_{ha}(t) \Pb[\widehat{\varphi}(\bZ) - \varphi(\bZ)|A=t] f(t) dt \\
        = &\,  \int g_{ha,j}(t)K_{ha}(t) \Pb \left \{ \left[ \left( 1-\frac{\rho(t,\bX)}{\widehat{\rho}(t,\bX)} \right)(\widehat{\mu}(t,\bX,1)-\mu(t,\bX,1))  \right] \widehat{w}(t,\bV)\bigg | A=t \right\} f(t) dt \\
        &\, + \int g_{ha,j}(t)K_{ha}(t) \Pb [(\tau(t,\bV) - \widehat{\tau}(t,\bV))(\widehat{w}(t,\bV)-w(t,\bV))|A=t] f(t) dt\\
        &\, + (\Pb_n - \Pb) \int g_{ha,j}(t)K_{ha}(t)  \widehat{\tau}(t,\bV)  f(t) dt
    \end{aligned}
    \]
    Here we slightly abuse the notation and denote $\Pb_n$ as the average over $D_2^n$. By boundedness of nuisance estimates and Cauchy-Schwarz's inequality, we have
    \[
    \begin{aligned}
        &\, \left |\int g_{ha,j}(t)K_{ha}(t) \Pb \left \{ \left[ \left( 1-\frac{\rho(t,\bX)}{\widehat{\rho}(t,\bX)} \right)(\widehat{\mu}(t,\bX,1)-\mu(t,\bX,1))  \right] \widehat{w}(t,\bV)\bigg | A=t \right\} f(t) dt \right|\\
        \lesssim &\, \int |g_{ha,j}(t)|K_{ha}(t)\|\widehat{\rho}-\rho\|_t\|\widehat{\mu}-\mu\|_t f(t) dt \\
        \leq &\, \sup_{|t-a|\leq h} \|\widehat{\rho}-\rho\|_t\|\widehat{\mu}-\mu\|_t \int \left| \frac{t-a}{h} \right|^{j-1}\frac{1}{h}K\left(\frac{t-a}{h} \right) f(t)dt \\
        = &\, \sup_{|t-a|\leq h} \|\widehat{\rho}-\rho\|_t\|\widehat{\mu}-\mu\|_t \int \left| u \right|^{j-1}K\left(u \right) f(a+hu)du \\
        \lesssim &\, \sup_{|t-a|\leq h} \|\widehat{\rho}-\rho\|_t\|\widehat{\mu}-\mu\|_t.
    \end{aligned}
    \]
    Similarly we have
    \[
    \begin{aligned}
        &\, \left | \int g_{ha,j}(t)K_{ha}(t) \Pb [(\tau(t,\bV) - \widehat{\tau}(t,\bV))(\widehat{w}(t,\bV)-w(t,\bV))|A=t] f(t) dt \right|\\
        \leq &\, \int | g_{ha,j}(t)|K_{ha}(t) \|\widehat{\tau}-\tau\|_t \|\widehat{w}-w\|_t f(t) dt\\
        \lesssim &\, \sup_{|t-a|\leq h} \|\widehat{\tau}-\tau\|_t\|\widehat{w}-w\|_t.
    \end{aligned}
    \]
    For the remaining term $ (\Pb_n - \Pb) \int g_{ha,j}(t)K_{ha}(t)  \widehat{\tau}(t,\bV)  f(t) dt$, by Lemma 2 in \citet{kennedy2020sharp} we have
    \[
    (\Pb_n - \Pb) \int g_{ha,j}(t)K_{ha}(t)  (\widehat{\tau}(t,\bV)-\tau(t,\bV))  f(t) dt = O_{\Pb} \left( \frac{\|\int g_{ha,j}(t)K_{ha}(t)(\widehat{\tau}(t,\bV)-\tau(t,\bV))f(t)dt \|_2 }{\sqrt{n}} \right)
    \]
    where
    \[
    \begin{aligned}
        &\,\left\|\int g_{ha,j}(t)K_{ha}(t)(\widehat{\tau}(t,\bV)-\tau(t,\bV))f(t)dt \right\|_2^2 \\
        =&\,  \int \left[ \int g_{ha,j}(t)K_{ha}(t)(\widehat{\tau}(t,\bV)-\tau(t,\bV))f(t)dt \right]^2 d\Pb(\bv) \\
        \leq &\,  \|\widehat{\tau}-\tau\|_{\infty}^2 \left(\int |g_{ha,j}(t)| K_{ha}(t) f(t)dt \right)^2 \\
        \lesssim &\,  \|\widehat{\tau}-\tau\|_{\infty}^2.
    \end{aligned}
    \]
    This together with $\|\widehat{\tau}-\tau\|_{\infty} = o_{\Pb}(1)$ implies 
    \begin{equation}\label{eq:empirical-term1}
    (\Pb_n - \Pb) \int g_{ha,j}(t)K_{ha}(t)  (\widehat{\tau}(t,\bV)-\tau(t,\bV))  f(t) dt  = o_{\Pb}\left( \frac{1}{\sqrt{n}}\right).
    \end{equation}
    By direct calculations (where we assume the outcome is bounded hence $\tau$ is also bounded)
    \[
    \begin{aligned}
        &\, \E \left\{\left[ (\Pb_n-\Pb)\int g_{ha,j}(t)K_{ha}(t)\tau(t,\bV)f(t)dt \right]^2 \right\} \\
        \leq &\, \frac{1}{n} \E\left \{ \left( \int g_{ha,j}(t)K_{ha}(t)\tau(t,\bV)f(t)dt \right) ^2\right\} \\
        \lesssim &\, \frac{1}{n} \E\left \{ \left( \int |g_{ha,j}(t)|K_{ha}(t)f(t)dt \right) ^2\right\}\\
        \lesssim &\, \frac{1}{n},
    \end{aligned}
    \]
    which implies
    \[
    (\Pb_n-\Pb)\int g_{ha,j}(t)K_{ha}(t)\tau(t,\bV)f(t)dt = O_{\Pb}\left( \frac{1}{\sqrt{n}}\right).
    \]
    Combining this equation with \eqref{eq:empirical-term1} yields
    \[
    (\Pb_n-\Pb)\int g_{ha,j}(t)K_{ha}(t)\widehat{\tau}(t,\bV)f(t)dt = O_{\Pb}\left( \frac{1}{\sqrt{n}}\right)= o_{\Pb}\left ( \frac{1}{\sqrt{nh}} \right).
    \]
    Hence under the rate conditions in the theorem, we conclude 
    \[
    R_2 = O_{\Pb}\left( \sup_{|t-a|\leq h} \|\widehat{\rho}-\rho\|_t \|\widehat{\mu}-\mu\|_t + \sup_{|t-a|\leq h} \|\widehat{\tau}-\tau\|_t \|\widehat{w}-w\|_t \right) + o_{\Pb}\left ( \frac{1}{\sqrt{nh}} \right) = o_{\Pb}\left ( \frac{1}{\sqrt{nh}} \right).
    \]
    The asymptotic normality then follows from Slutsky's theorem.
\end{proof}

\section{Additional Simulation Results}

\subsection{Nuisance Functions Estimated by Parametric Methods}\label{appendix:simu_parametric}
\begin{figure}[ht]
    \centering
    \includegraphics[scale=0.75]{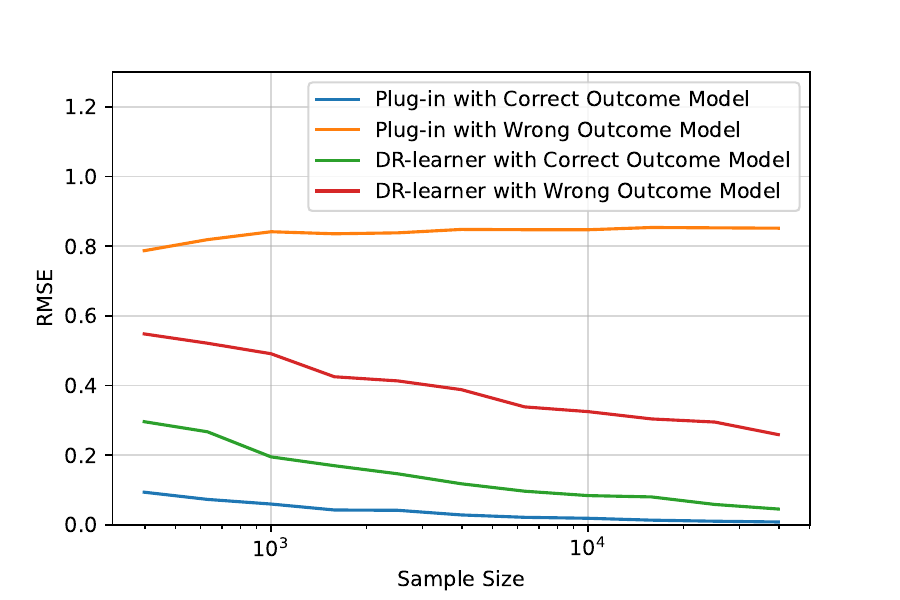}
    \caption{RMSE versus sample size (in log scale) when nuisance functions are estimated by parametric models.}
    \label{fig:rmse-parametric}
\end{figure}

We evaluate the performance of plug-in-style estimator and doubly robust estimator when nuisance functions are estimated using parametric models under the same setting as Section \ref{sec:simulation}. We will fit linear regression models for $\mu, \tau, \lambda$ and separately consider correctly specifying the outcome model $\mu, \tau$ or not, where a misspecified model left out the quadratic term in $a$ but keeps all other main effects and interactions. The conditional density of $A$ given $\bV$ is obtained by first estimating the conditional mean of $\E[A|\bV]$ and plug-in the normal density (i.e., we assume the model for conditional density is always correct). For the plug-in estimator we randomly separate the sample into two parts $D,T$. The first part $D$ is used to fit the regression models for all nuisance functions and we take the average on the second part as $\widehat{\theta}(a) = \frac{1}{|T|}\sum_{i \in T} \widehat{\tau}(a, \bV_i)$. The roles of $D,T$ are then exchanged to obtain another estimate and the final estimator is the average of two estimates. The doubly robust estimator is implemented according to Algorithm \ref{alg:DR-surrogates}, where the bandwidth is selected on $D_2$. We generate samples with sample size $n \in \{ 10^{2.6}, 10^{2.8}, \dots, 10^{4.6}\}$, apply two estimators to estimate ${\theta}(1)$ and repeat the process $500$ times. The results are summarized in Figure \ref{fig:rmse-parametric}.

When the outcome model is misspecified, the plug-in estimator (solely based on outcome modeling) is no longer consistent, as shown in Figure \ref{fig:rmse-parametric}. The doubly robust estimator, however, models both the outcome and treatment process and has a smaller estimation error when the outcome model is misspecified. Estimation with correctly specified outcome model corresponds to $\alpha = 0.5$ in Figure \ref{fig:rmse_alpha}, where doubly robust estimator with outcome model and propensity score both correctly specified has larger error compared with the plug-in estimator since slow rate of local linear smoothing $O(n^{-2/5})$ dominates the nuisance estimation error $O(n^{-1/2})$. 

\subsection{Nuisance Functions Estimated by Nonparametric Methods}\label{appendix:simu_nonparametric}

We further evaluate the performance of the plug-in style estimator and doubly robust estimator when nuisance functions are estimated using nonparametric models under the same setting as Section \ref{sec:simulation}. We will fit nuisance functions $\mu, \tau, \rho$ by superlearner combining generalized linear models and random forests. The conditional density of $A$ given $\bV$ is estimated by kernel density estimation. The plug-in estimator is implemented the same way as in Appendix \ref{appendix:simu_parametric} and the doubly robust estimator is implemented according to Algorithm \ref{alg:DR-surrogates}. We generate samples with sample size $n \in \{ 10^{2.6}, 10^{2.8}, \dots, 10^{4}\}$, apply two estimators to estimate ${\theta}(1)$ and repeat the process $500$ times. The results are summarized in Figure \ref{fig:rmse-parametric}.
\begin{figure}[ht]
    \centering
    \includegraphics[scale=0.75]{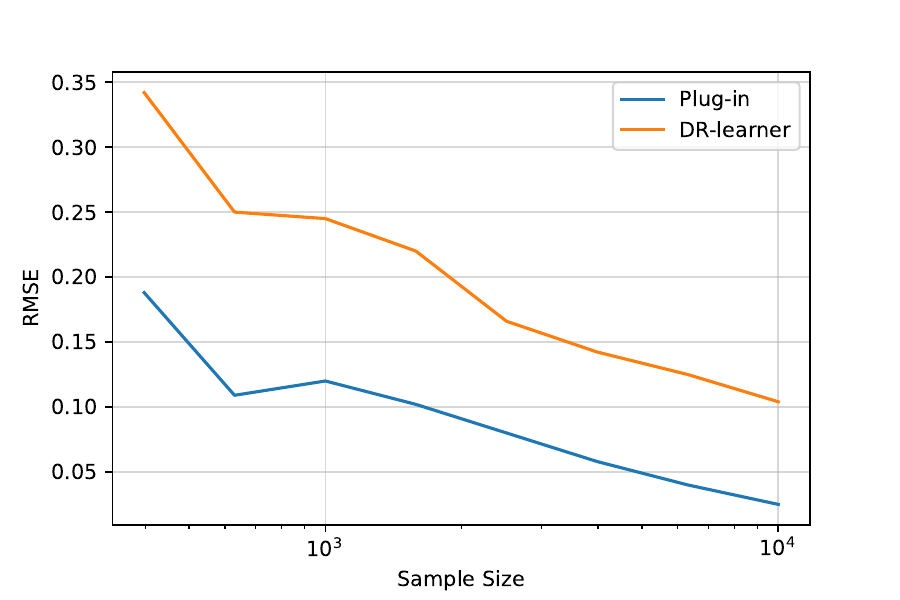}
    \caption{RMSE versus sample size when nuisance functions are estimated by nonparametric models.}
    \label{fig:rmse-nonparametric}
\end{figure}

As shown in Figure \ref{fig:rmse-nonparametric}, the plug-in-style estimator has smaller estimation error when all the nuisance functions are fitted by nonparametric methods. This could be explained by the diffculty in nonparametrically estimating conditional density $\pi$. Due to the curse of dimensionality, a large sample size is required for the kernel density estimator to estimate $\pi$ well. In our simulations, we find $\widehat{\pi}$ could be small and hence violate the positivity assumption, yielding variation in the construction of pseudo-outcome $\widehat{\varphi}$ and larger estimation error of dose-response compared with the plug-in estimator. In applications, prior knowledge on the conditional density, including a reliable parametric model on $\pi$ from domain knowledge or information on the lower bound on $\pi$ due to study design might help us reduce the variation of $\widehat{\pi}$ and estimate the conditional density better, which could improve the performance of doubly robust estimator proposed.

\subsection{Surrogates Dependent on Treatment and Covariates}

\begin{figure*}[ht]
	\centering
	\subfigure[$n=500$]{
		\begin{minipage}[t]{0.45\linewidth}
			\centering
			\includegraphics[width=3in]{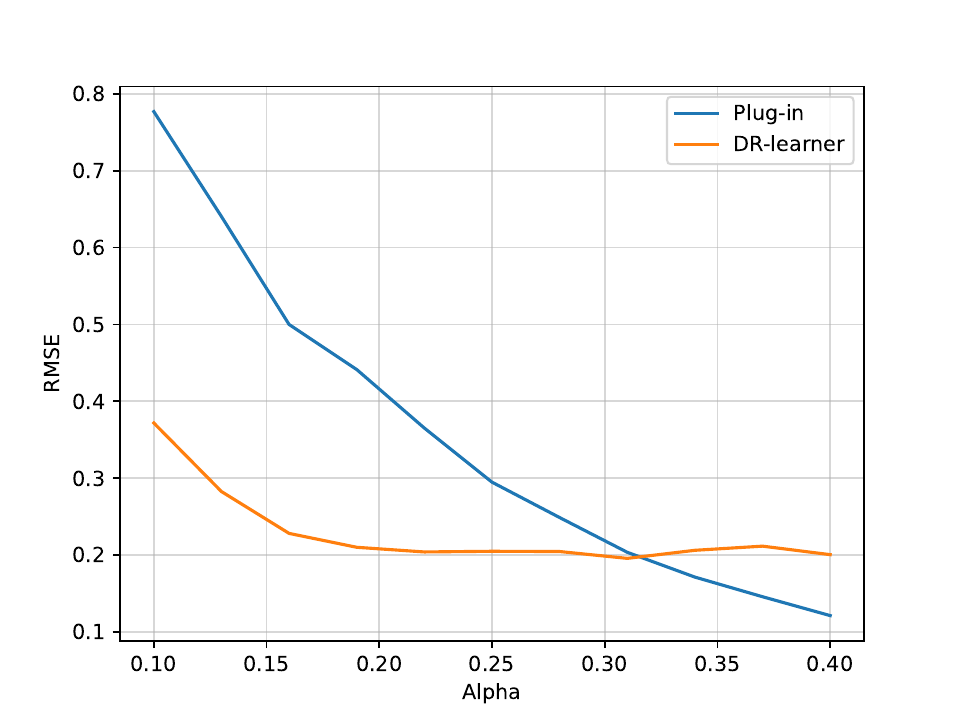}
	\end{minipage}}
	\subfigure[$n=2000$]{
		\begin{minipage}[t]{0.45\linewidth}
			\centering
			\includegraphics[width=3in]{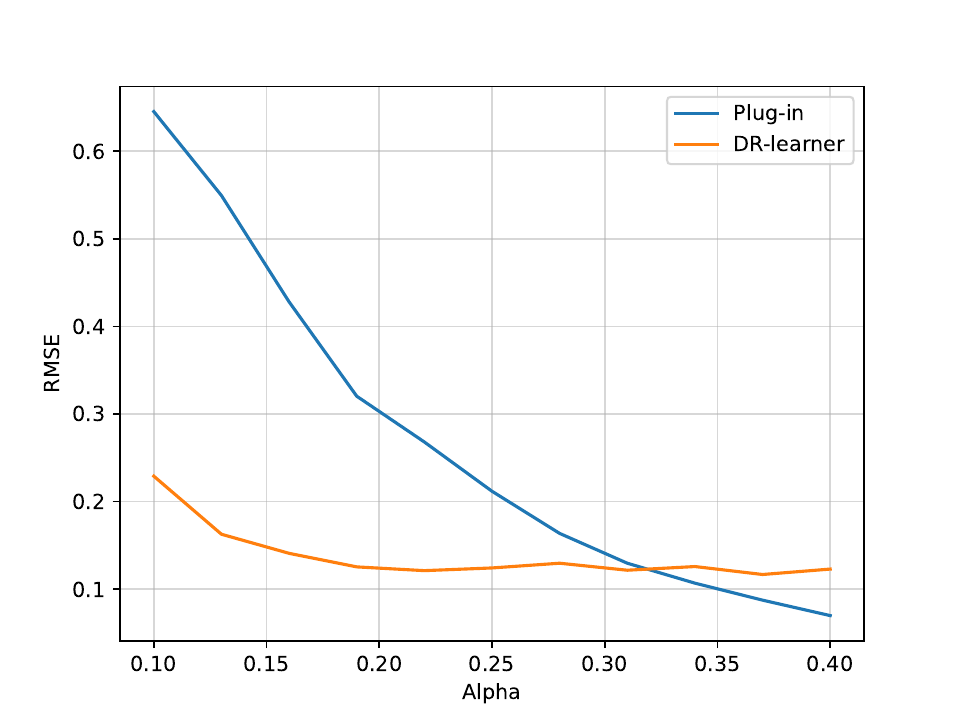}
	\end{minipage}}
	\centering
	\caption{Root mean square error Versus $\alpha$, where $n^{-\alpha}$ is the estimation error of the nuisance functions.}
	\label{fig:rmse_alpha_depend}
\end{figure*}

In this section we provide additional simulation results on the setting where surrogates may depend on the treatment and covariates. In real applications, the surrogates $\bS$ are post-treatment short-term outcomes and are likely to depend on pre-treatment covariates and the treatment. We follow the same setting and estimation procedures as in Section \ref{sec:simulation} (so the nuisance estimation error is set manually) but modify the conditional distribution of $\bS$ given $(A,\bV)$ as 
\[
\bS \sim N\left((V_1+A,V_2-A)^{\top}, \boldsymbol{I}_2\right).
\]
The RMSE is estimated from $M=500$ replications of the data-generating process and estimation procedures. The results of DR-learner and plug-in estimators are summarized in Figure \ref{fig:rmse_alpha_depend}.

Figure \ref{fig:rmse_alpha_depend} shows when surrogate outcomes $\bS$ depend on treatment and covariates, the doubly robust estimator enjoys better estimation accuracy compared with the plug-in estimator when $\alpha$ is small (and hence nuisance estimation error is large). As $\alpha$ increases and the nuisance estimation error becomes smaller, the plug-in estimator gradually outperforms the doubly robust estimator. The results are similar to those in Section \ref{sec:simulation} and readers are referred to Section \ref{sec:simulation} for more discussion and explanation.

\subsection{Comparison Between Supervised and Semi-supervised Methods}

\begin{figure}[ht]
    \centering
    \includegraphics[scale=0.75]{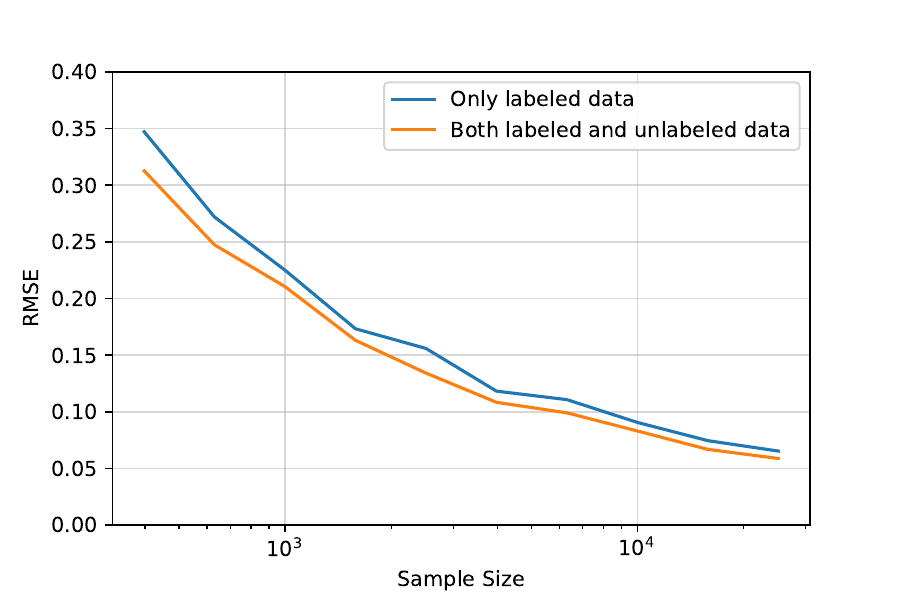}
    \caption{RMSE versus sample size when nuisance functions are estimated by correctly specified parametric models.}
    \label{fig:rmse-compare}
\end{figure}

In this section we compare the supervised estimator, which implements the method in \cite{kennedy2017non} on labeled data $\mathcal{L}$, with our semi-supervised estimator incorporating the unlabeled data $\mathcal{U}$ with surrogate outcomes. We follow the same setting in Section \ref{sec:simulation} and implement method in \cite{kennedy2017non} via similar cross-fitting techniques as in Algorithm \ref{alg:DR-surrogates} except that there is no need to fit $\rho(a,\bx)= \Pb(R=1 \mid A=a, \bX = \bx) $ since their method only uses labeled data. All the nuisance functions are estimated by correctly specified parametric models. We generate samples with sample size $n \in \{ 10^{2.6}, 10^{2.8}, \dots, 10^{4.6}\}$, apply supervised and semi-supervised methods to estimate ${\theta}(1)$ and repeat the process $M=500$ times. The results are summarized in Figure \ref{fig:rmse-compare}.

Figure \ref{fig:rmse-compare} shows the semi-supervised method incorporating unlabeled data with surrogate outcomes has a smaller estimation error compared with the supervised method only using labeled data. The improvement comes from many aspects: As discussed in Section \ref{sec:DR-normality}, the asymptotic variance of our semi-supervised method is smaller than that of the supervised method only using labeled data; Also in the semi-supervised setting, both labeled and unlabeled data are used to estimate the nuisance functions, making nuisance estimation more accurate than supervised method solely based on labeled data since the effective sample size of semi-supervised method is larger.

\section{Real Data Analysis}\label{appendix:real-data}

In this section we apply the proposed method to the Job Corps study, conducted in the 1990s to evaluate the effects of the publicly funded U.S. Job Corps program. The Job Corps program targets a population between 16 and 24 years old living in the U.S. and coming from low-income households, where participants received an average of 1,200 hours of vocational training over approximately eight months. \citet{schochet2001national} and \citet{schochet2008does} discuss the study design in detail and analyze the effects of the Jobs Corps program on various outcomes. They found the Job Corps program effectively increased educational attainment, prevented arrests, and increased employment and earnings. The effects of the Job Corps program have been extensively studied under different causal inference frameworks \citep{flores2009identification, huber2014identifying, frolich2017direct}. 

However, these previous studies on the Job Corps program mainly considered binary treatment definitions. In this work, we are interested in estimating the effects of different doses of participation in the program on future involvement in the criminal justice system, namely the number of arrests in the fourth year after the program (outcome $Y$). Specifically, our treatment variable $A$ is defined as the total hours spent either in academic or vocational classes of the program. The short-term surrogate outcome $S$ is the proportion of weeks employed in the second year after the program. \citet{huber2020direct} used generalized propensity score weighting to estimate the continuous treatment effects of time spent in the Job Corps program under a mediation analysis framework. We re-analyze their dataset publically available on Harvard dataverse \citep{DVNAJ7Q9B2020}, using the doubly robust estimator proposed as an illustration of our method. 

\begin{figure}
    \centering
\includegraphics[scale=0.75]{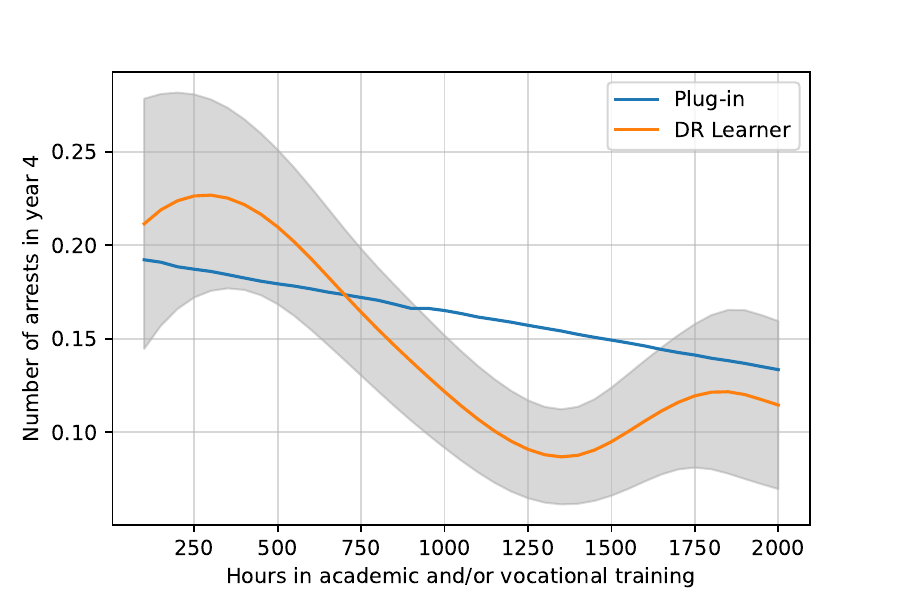}
    \caption{Dose response curve of number of arrests in year 4 (outcome) versus hours in academic and/or vocational training (treatment)}
    \label{fig:real-data}
\end{figure}

We follow \citet{huber2020direct} and focus on $n=4000$ samples with a positive treatment (i.e., $A_i>0, 1 \leq i \leq n$). To identify the causal estimand, we invoke the conditional exchangeability in Section \ref{sec:identification}, where a set of covariates is conditioned on to adjust for confounding bias. The covariate set $\bV$ we adjust for confounding consists of age, gender, ethnicity, education, marital status, previous employment status and income, welfare receipt during childhood, and family background (e.g., parents’ education). Missing dummies are created for covariates in $\bV$ containing missing values. The readers are referred to Table 4 in \citet{huber2020direct} for descriptive statistics of the pretreatment covariates as well as the treatment, surrogate, and outcome
variables in the data. Conditioning on a rich set of covariates is important since it enables us to identify the causal estimand by making the conditional exchangeability assumption plausible. In our analysis, the outcome $Y $ (number of arrests in year 4) is omitted for 25\% of the samples randomly to mimic the setting where the primary outcome is missing and a surrogate outcome is used as auxiliary information. We estimate the treatment effects $\theta(a)$ for each of $a \in \{100,150,200,\dots,2000\}$ using both plug-in-style estimator and doubly robust estimator in Algorithm \ref{alg:DR-surrogates}, where the nuisance functions $\rho, \mu, \tau$ are estimated by superlearner \citep{van2007super} combining generalized linear model and random forests, conditional density $\pi$ is estimated by kernel density estimator. The estimated dose response curve is plotted in Figure \ref{fig:real-data}.

Figure \ref{fig:real-data} shows that, as participants spend more hours in the program, the expected number of arrests in year 4 has a decreasing trend, which confirms the conclusion that such training programs effectively reduce involvement in the criminal justice system. Importantly, the dose-response fitted by the doubly robust estimator is very similar to the results in \citet{huber2020direct}: the shape of the function is similar to their Figure 2. (Note that the specific values on the y-axis are different since they plot a contrast effect and we instead plot the expectation of potential outcome $Y^a$.) As pointed out in \citet{huber2020direct}, the treatment effect is highly nonlinear, which is further verified by the curve estimated from the doubly robust estimator. By contrast, the curve estimated by a simple plug-in estimator in \eqref{eq:plugin} fails to capture this non-linearity, possibly because it suffers from a large first-order bias.

\end{document}